\documentclass[graybox]{svmult}

\usepackage{type1cm}        
\usepackage{makeidx}         
\usepackage{graphicx}        
\usepackage{threeparttable}                             
\usepackage{tikz}
\usepackage{multicol}        
\usepackage[bottom]{footmisc}
\usepackage{wrapfig, lipsum}
\usepackage{caption}
\usepackage{float}
\usepackage{subcaption}
\usepackage{eqnarray}
\usepackage{tcolorbox}
\usepackage{hyperref}
\usepackage{newtxtext}       %
\usepackage{newtxmath}       
\usepackage{gensymb}
\usepackage[export]{adjustbox}
\usepackage[backend=bibtex8,sorting=none]{biblatex}
\usepackage{array} 
\providecommand{\authormark}[1]{$^{#1}$}
\usepackage{amsmath}
\usepackage{newtxtext,newtxmath}
\makeindex             

\begin{document}

\title*{Where to Perch in a Tree: Vision-Guidance for Tree-Grasping Drones}

\author{A Dunnett\authormark{1}, L Bottomley\authormark{1}, M Kovac\authormark{3} and BB Kocer\authormark{1}}

\institute{1 \at Department of Civil, Aerospace and Design Engineering, University of Bristol.
\and 2 \at Laboratory of Sustainability Robotics at Swiss Federal Laboratories for Materials Science and Technology (EMPA); and  Ecole Polytechnique Fédérale de Lausanne (EPFL).}

\maketitle

%
%

\abstract{
This study demonstrates a method to locate an ideal perch location on a tree for vision-guided autonomous tree-perching drones. Various image processing algorithms, including those used for machine learning image segmentation and binary image morphology, are implemented to assess the shape and structure of a tree. Rather than identifying the closest available branch, this study builds on vision methods by evaluating the potential of each branch, determining its suitability for perching based on factors such as branch width, slope (angle to the horizontal) and curvature. For a given tree-perching drone and a dataset of more than 10,000 urban tree images taken from February to October in a subtropical and temperate monsoon climate, the proposed method successfully produces a result for $76\%$ of feasible targets. A feasible target defined as a tree where the branch diameters are sufficiently thick and where the available perching space is at least equal to the width of a tendon-driven grasping claw. These successful preliminary results create a foundation from which a number of identified improvements and additional features can be developed to create a generalised method; this will involve the incorporation of supplementary data from depth perception and attitude sensors to enhance the branch assessment.}

\begin{figure}[b] 
    \centering
    \includegraphics[width=7.8cm]
    {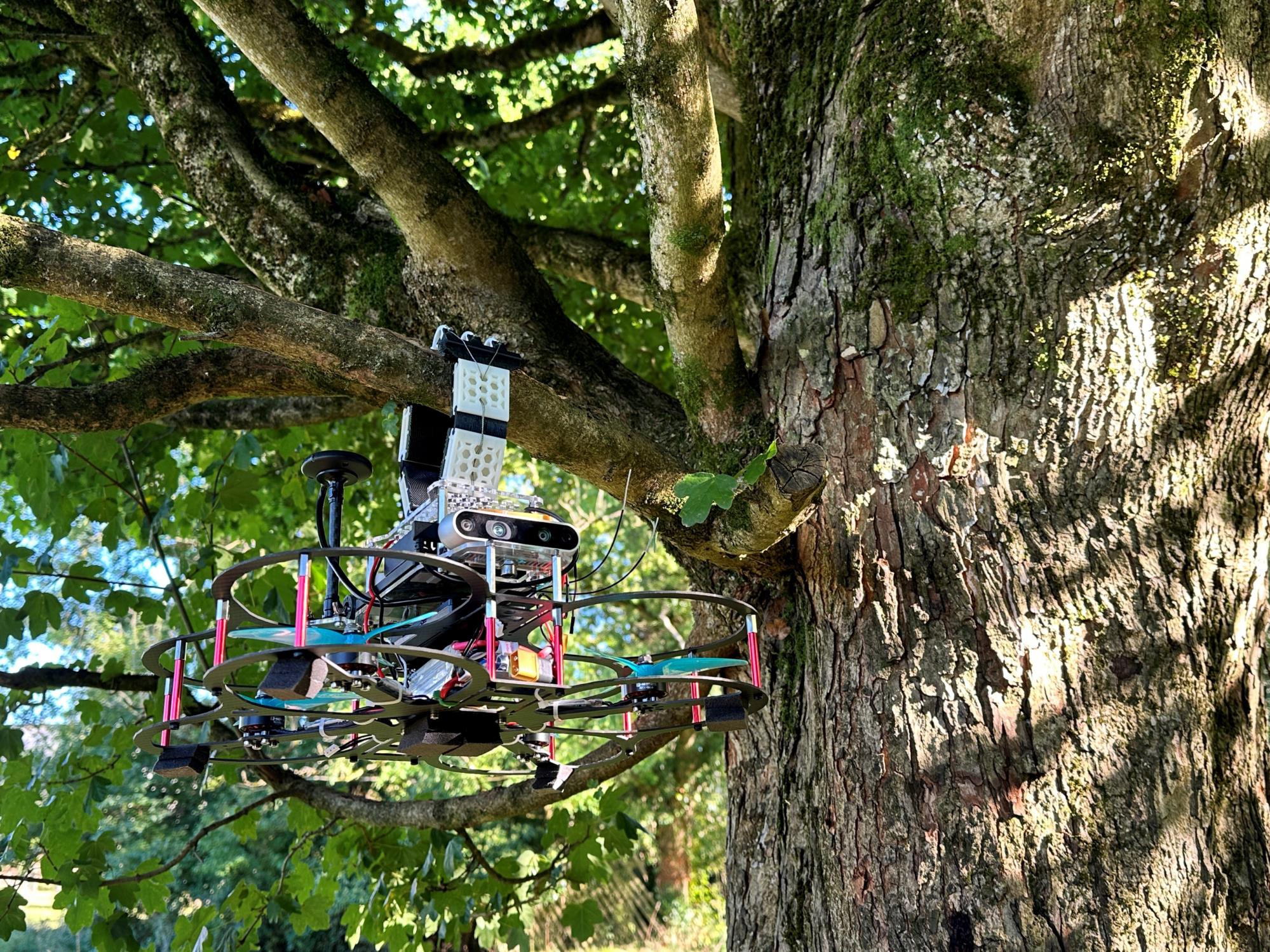}
    \caption{Tendon-driven grasper design perching on tree branches.}
    \label{tree-grasping-drone}
\end{figure}
    
\newpage

\section{Introduction}
\subsection{Context and Motivation}
\label{sec:1}

The surge in technologies enhanced with Artificial Intelligence (AI) also brings with it a significant environmental footprint, mainly due to the energy required during model training and hardware manufacturing processes \cite{wu_sustainable_nodate}. The most recent report by the United Nations on the status of Sustainable Development Goal number 15, Life on Land, considers the threat to biodiversity and forestry to be one of the primary global challenges facing humanity \cite{united_nations_department_for_economic_and_social_sustainable_2024}.

Aerial robots, commonly referred to as drones, could be considered as flying, physical manifestations of AI, and represent one avenue through which AI technologies might support efforts to mitigate or even reverse the ongoing decline in biodiversity. For example, there are multiple examples in the literature of drones being used to deploy sensors, \cite{geckelerBistableHelicalOrigami2022} take images \cite{zhangSeeingForestDrones2016c,hoVisionBasedCrown2022, batesLeafLevelAsh2025} or collect physical samples \cite{hamazaDesignModelingControl2020a, kocerForestDronesEnvironmental2021a} in forest environments for the purposes of conservation. To achieve this objective, researchers have developed a multitude of multi-rotor drones to aid in forest conservation by navigating densely forested environments to retrieve, in a minimally invasive manner, field data samples \cite{xiao2021optic,kocer2022immersive,pringleRoboticsAutonomousSystems2023,romanello2024exploring}.

However, problems remain with this approach. One significant issue is that the onboard energy storage capabilities of drones can limit the duration of environmental sensing missions. One possible solution is perching, a process by which a drone could, analogously to a bird, rest from flight by grasping a tree branch. By perching, a drone could continue to actively sense or execute other functions while expending energy at a lower rate. As a result, mission durations could be significantly extended and, therefore, the number of data gathering actions, such as sampling or detection, per mission could be increased \cite{li_tendon-driven_2025,zhengMetamorphicAerialRobot2023a,zuffereyHowOrnithoptersCan2022,kovacPerchingMechanismMicro2009}.

Developing the capability for drones to successfully perch in unstructured environments requires overcoming a series of inter-disciplinary hurdles. One such hurdle that this work will aim to overcome is how to identify a location for perching to occur considering that, in a forest environment, access to external localisation methods, such as GNSS, are restricted \cite{pritchard2025forestvo}. Therefore, an onboard vision-based method for analysing trees in real time, harnessing various image processing algorithms to identify the optimal perching location, is crucial.

An existing method to identify the perch site was developed by Li et al. \cite{li_tendon-driven_2025}. Their method was deployed onboard a quadrotor drone that had been equipped with a tendon-driven claw perching mechanism, as shown in Figure \ref{tree-grasping-drone}.  To achieve this, Li et al. used the DeepLabV3+ architecture for semantic image segmentation, implemented on PyTorch, for branch and trunk detection. They employed a ResNet-34 encoder with pre-trained weights as the backbone. Their PLI method first detects the trunk of a tree in an RGB-D image and, from there, aims to identify the centre of the branch considered to be nearest.

To improve upon the existing method, the main objective of this research is to develop a decision-making algorithm capable of selecting the optimal branch based on a defined set of criteria.

Despite no research existing, to the best of our knowledge, for this specific use case, studies from other disciplines were identified that utilised frameworks that could be applied for this application.
The FilFinder \cite{koch_github_2025} algorithm presents a complete image processing and analysis process to characterise the structure of galaxy filaments, which resemble the skeletal structure of a tree. This approach also presents an interesting solution for characterising the orientations of the structure, which is a key element of the suitability of a tree branch for perching. Other similar algorithms in various research domains follow the process of analysing a skeletal structure to create an understanding of the original structure; for example, in the medical field, the skeletal structure of a neuron is analysed \cite{nunez-iglesias_new_2018} to understand how malaria alters the structure of the red blood cell.

Critical to the success of this task will be to ensure that parameters of interest to the perching PLI method can be extracted from an image of a tree in a cluttered and uncontrolled environment. Birds are exceptionally good at this, identifying a suitable branch from hundreds of options, for perching in real-time, while flying at high velocities. The question is, how can a perching drone achieve this same incredible feat?

\subsection{Objectives}\label{objectives}

This research aims to create a PLI methodology to improve the perching capabilities of drones. To achieve this, we aim to develop an image processing pipeline that extracts information from an image of a tree to create a graph-based representation of the tree that only contains feature information relevant to the application. By defining a set of criteria that relate the optimal angle, width, and curvature of the 'ideal' branch set by the specifications of the tree-grasping drone shown in Figure \ref{tree-grasping-drone} the graph-based tree can be analysed to output the perching location for any tree image provided. In this work we define success such that the PLI method should produce an ideal perching location for at least 4 out of 5 inputs (or 80\%).

Furthermore, the PLI method should be structured in a way that provides a foundation for future developments to be incorporated in a modular fashion. For example, to accommodate the identification of possible obstacles around the branches. Finally, a quantitative survey of the processing time and computational expense of the method should be conducted. This is because, firstly, reducing the computation time of the PLI method will increase the frequency at which the target position is updated, increasing the capability of the moving drone to perform accurate state estimation in real-time. Secondly, as the method will be executed on an embedded platform with severe onboard energy constraints, reducing the computational resources required will increase the maximum mission duration.

\begin{figure}
    \centering
    \includegraphics[width=0.75\linewidth]{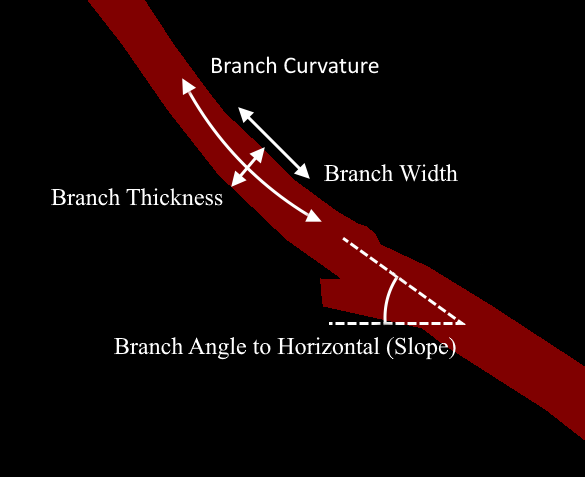}
    \caption{Annotated key features on segmented tree mask.}
    \label{fig:tree_mask_annotation}
\end{figure}

\section{Methodology}
\subsection{Process Overview}


The steps involved in the study can be simplified into 3 main tasks and are illustrated in Figure \ref{fig:tree_mask_annotation}: 
\begin{enumerate}
    \item Instance segmentation (Section \ref{step1})
    \item Feature extraction and data preparation (Section \ref{step2})
    \item Profiling and analysis (Section \ref{step3})
\end{enumerate}

\noindent
Each stage of the process is broken down into individual tasks. This is shown in Figure \ref{fig:master-flowchart} (a). 

\subsection{Dataset Suitability}\label{app:dataset}

To train and validate the PLI method described in this work, the Urban Tree date set by Yang et al. was used \cite{BranchSegmentation_2022}. This dataset is composed of 41,467 high-resolution classification images (22,872 with annotated images) that covered 50 tree species that were captured in a variety of conditions. The founding purpose of the dataset was to aid in the task of training segmentation models for the purpose of tree species identification. However, we propose that the same segmentation model would also be applicable for the purposes of our PLI method. However, we acknowledge limitations of the dataset in the following ways.

\begin{itemize}
    \item This dataset is comprised exclusively of deciduous trees that have shed their foliage. Though this is useful from a segmentation perspective (the boundary between the tree and the background is very defined), the drones are intended to obtain samples from the tree and, as such, should be able to perch on a tree with full foliage. 
    \item The majority of the images contain the whole tree, which for this study is necessary due to the method used to define the pixel to length ratio. However, in practice, it may be necessary to profile the tree using a closer image that contains only a section of the entire tree. For example, in dense forest, where there is limited space between trees.
    \item The camera used to produce the images in the dataset will likely have a sensor resolution and lens focal length that differ from the resolution and focal length of the drone mounted camera. As a result, the images analysed onboard the drone will be distinct from the dataset images. In practice, the PLI method has been able to overcome these differences. However, in future, producing a dataset from images gathered by the relevant drone could improve accuracy.
    \item Once the test environment has been determined for the drone, it would be beneficial to include tree species that are geographically native to the test region; increasing the similarities between the training dataset and real-world subjects.
\end{itemize}

 Despite the limitations, we have deemed the Urban Tree dataset to be the most appropriate choice for our work as it remains the only publicly available and fully annotated dataset of living trees taken from a below-canopy perspective.

\subsection{Image Segmentation}\label{step1}

To create a binary mask representation of the tree structure, a pre-existing semantic segmentation model must be trained using a representative dataset. By using transfer learning in this way we can lower computational costs, require smaller datasets during training and, ultimately, maintain an accurate and well-generalised model. \cite{murel_what_2024} 

\begin{figure}[H]
    \centering
    \includegraphics[width=1.0\linewidth]{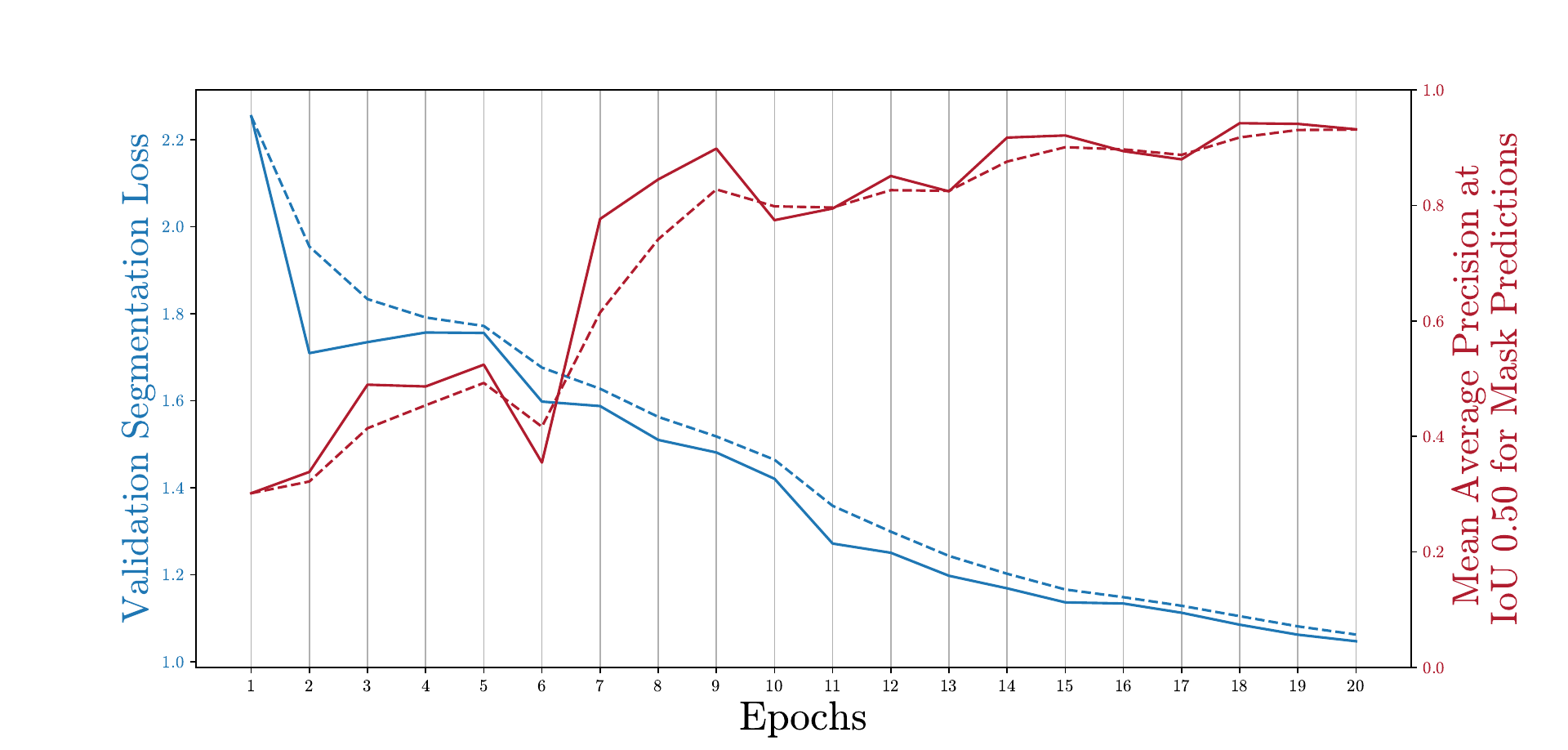}
    \caption{Training loss (left) and Segmentation Precision (right) for Ultralytics YOLOv11 Model.}
    \label{fig:model_training}
\end{figure}

For this study, Ultralytics YOLOv11 \cite{ultralytics_inc_ultralytics_2024} was used as the base model, and so the dataset \cite{BranchSegmentation_2022} was converted from a binary mask image format (see Appendix \ref{app:more_info}, Figure \ref{fig:mask}) to the required format for model training. Due to the well-designed implementation of the Ultralytics framework, once all training data have been formatted and specified, the training process can be performed with no user intervention; the training parameters are set and optimised during training automatically. The model should maintain a steady reduction in loss and an increase in validation accuracy (see Figure \ref{fig:model_training}). The loss value directly impacts the training parameters (convolutional layer weights and learning rate, for example) after each epoch. Mean Average Precision at Intersection over Union (IoU) of 50\% for Mask Predictions (Figure \ref{fig:model_training}, right axes) is the precision of the model at a 50\% mask overlap threshold. This lower overlap is shown as it is not preferred to obtain 100\% precision for complete overlap, as this indicates the model has completed too many epochs and has become over-fit.

\begin{figure}[H] 
    \centering
    \begin{minipage}{0.5\textwidth}
        \centering
        \includegraphics[width=2\linewidth]{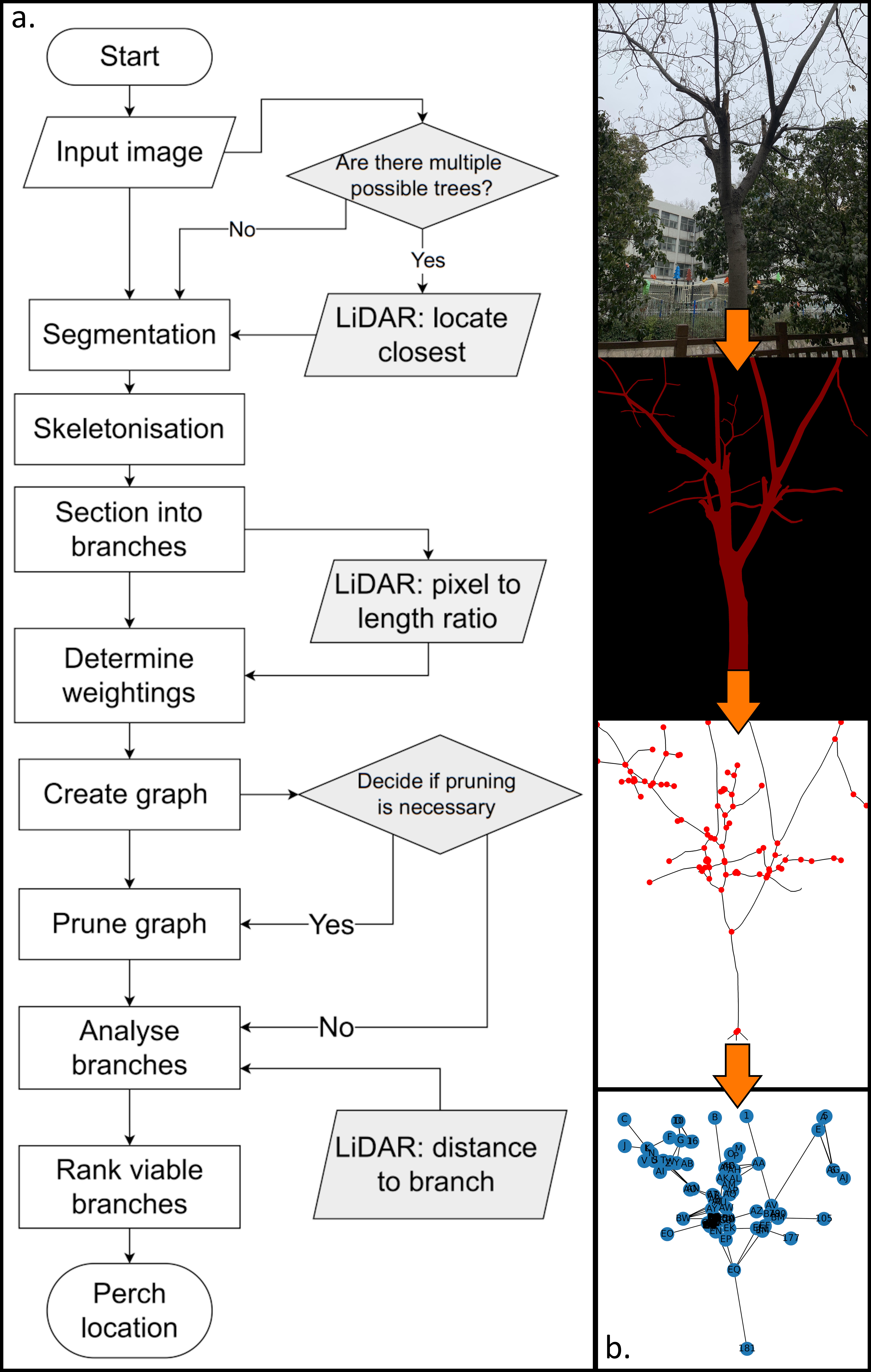}
        \label{fig:flowchart}
    \end{minipage}%
    \hfill
    \begin{minipage}{0.5\textwidth}
        \centering
            

    \end{minipage}
    \label{fig:img_flowchart}
    \captionsetup{justification=centering}
    \caption{Process flowchart (a, left), where possible incorporation of additional data (see \ref{future-dev}) to improve process robustness is listed alongside the main pipeline. Example key visualisations of the data process throughout (b, right).}
    \label{fig:master-flowchart}
\end{figure}

\subsection{Feature Extraction and Data preparation}\label{step2}

\subsubsection{Skeletonisation}
Once the binary segmentation mask has been extracted, this must be altered into a form in which the local angles, widths, and shape of each branch can be determined.
Through the use of the erosion operation, the mask can be reduced down to a single, pixel wide, 'skeleton'. Furthermore, the Medial Axis Transformation (MAT) also returns the nearest background pixel at any point on the skeleton. Thus, this is approximately the width of the structure at each skeleton point. This is shown in Figure \ref{fig:MAT} for an example tree (see Appendix \ref{app:more_info}, Figure \ref{fig:orig_img}).

\begin{figure}[H]
    \centering
    \includegraphics[trim= 0cm 0cm 0cm 0.7cm ,clip, width=0.5\linewidth]{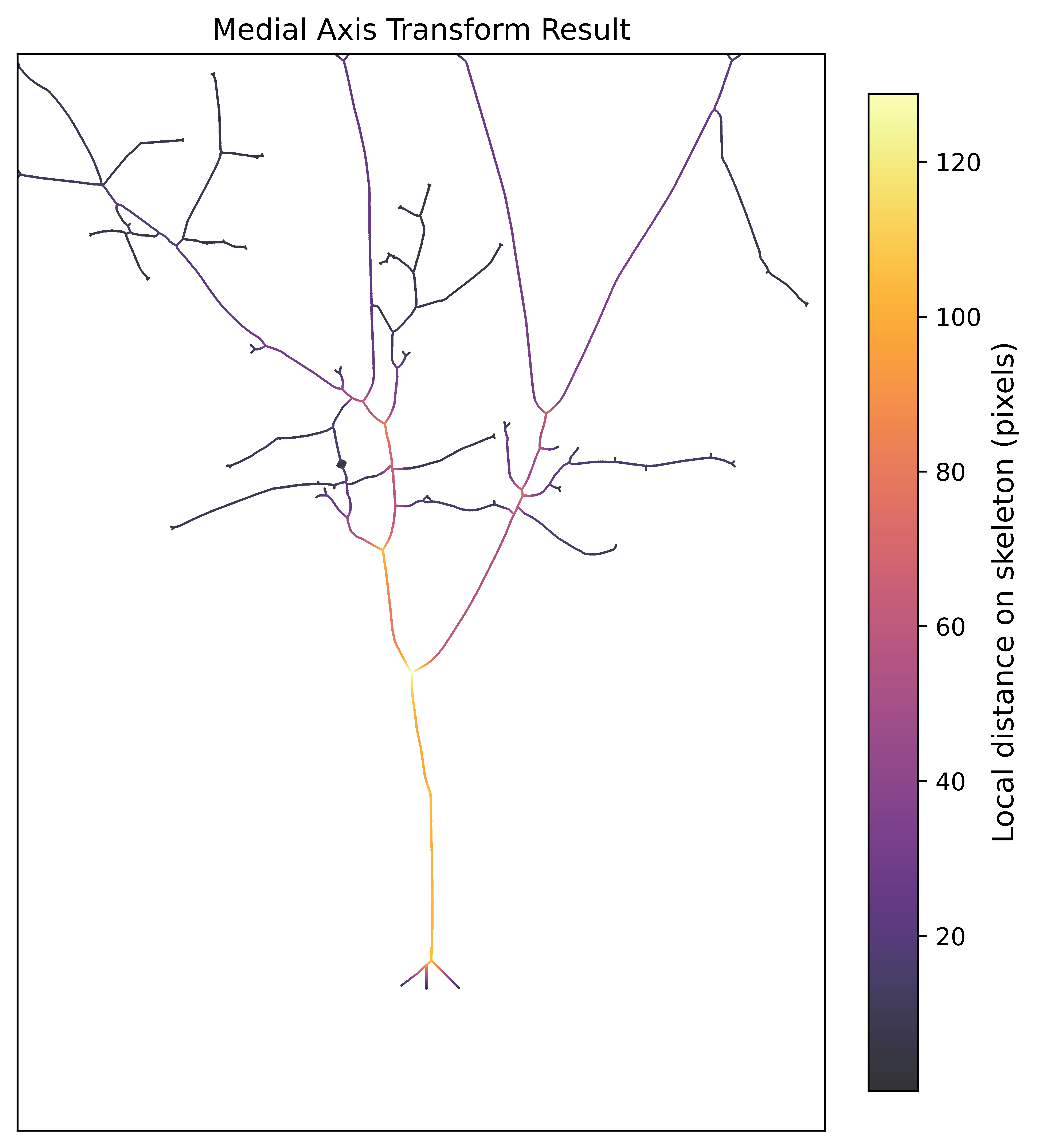}
    \caption{\centering Medial Axis Transform for Example Tree depicting the pixel coordinates of the skeleton and the corresponding thickness of the original structure at each point.}
    \label{fig:MAT}
\end{figure}

\subsubsection{Section into Branches}

From this skeleton, the intersection points (or 'nodes') of the structure should be identified to begin sectioning the tree. An algorithm ('\texttt{Find\_FilPix}') from the FilFinder package \cite{koch_github_2025} was adapted for this functionality.
This algorithm scans the skeleton using a 3x3 pixel kernel to classify each pixel into one of the options shown in Table \ref{tab:types_of_points}. It classifies the intersections assuming 8-point connectivity (see Appendix \ref{app:imgs}, Figure \ref{fig:connectivity}), where pixels located diagonally from the reference (centre of the kernel) are considered connected as one object.

\begin{table}[!t]
\centering
\begin{threeparttable}
\caption{Types of Points Identified from the Skeleton}
\label{tab:types_of_points}       
%
%
\begin{tabular}{p{4cm}p{4cm}}
\hline\noalign{\smallskip}
Type of Point & Number of Neighboring Pixels\\
\noalign{\smallskip}\svhline\noalign{\smallskip}
End points & 1\\
Body points & 2  \\
T-junction points & 3  \\
Block/intersection points & 4 or more\\
\noalign{\smallskip}\hline\noalign{\smallskip}
\end{tabular}
    \end{threeparttable}
\end{table}

{By removing the pixels that are identified as intersections from the original skeleton, the tree structure can then be sectioned into individual branches. The 'label' method from the package 'Scipy.ndimage' \cite{noauthor_multidimensional_nodate} is used to identify objects within a binary image. In order to group objects with 8-point connectivity (required here where the structure is one pixel wide), the scanning kernel used is a 3x3 ones matrix. The data are returned in a format where each pixel takes a label from 1 through to \textit{n} (where \textit{n} is the total number of objects), and this branch labelling system persists throughout the pipeline. 
\par
At this point the data can be considered as an array of pixel values, where the background is 0 (or False) and the pixels containing a branch take the label value. To prepare the data for the subsequent processing steps, these branches should be extracted to arrays containing each (\textit{x},\textit{y}) co-ordinate in its correct order. An exploration of the methods to format the data in this way is included in Appendix \ref{info:branch_ordering}. 
}

\subsubsection{Create Graph}

At this stage in the image processing pipeline, the extracted branch data are sufficient to calculate the basic criteria; branch width, local curvature, and angle to the horizontal axis. However, further processing is required to gain a full understanding of the tree's structure and the inter-connectivity of its branches, allowing for future developments that take this information into account at the analysis and decision stages. 

\begin{figure}[h] 
    \centering
    \includegraphics[trim=1cm 2.5cm 1cm 3.3cm, clip, width=\linewidth]{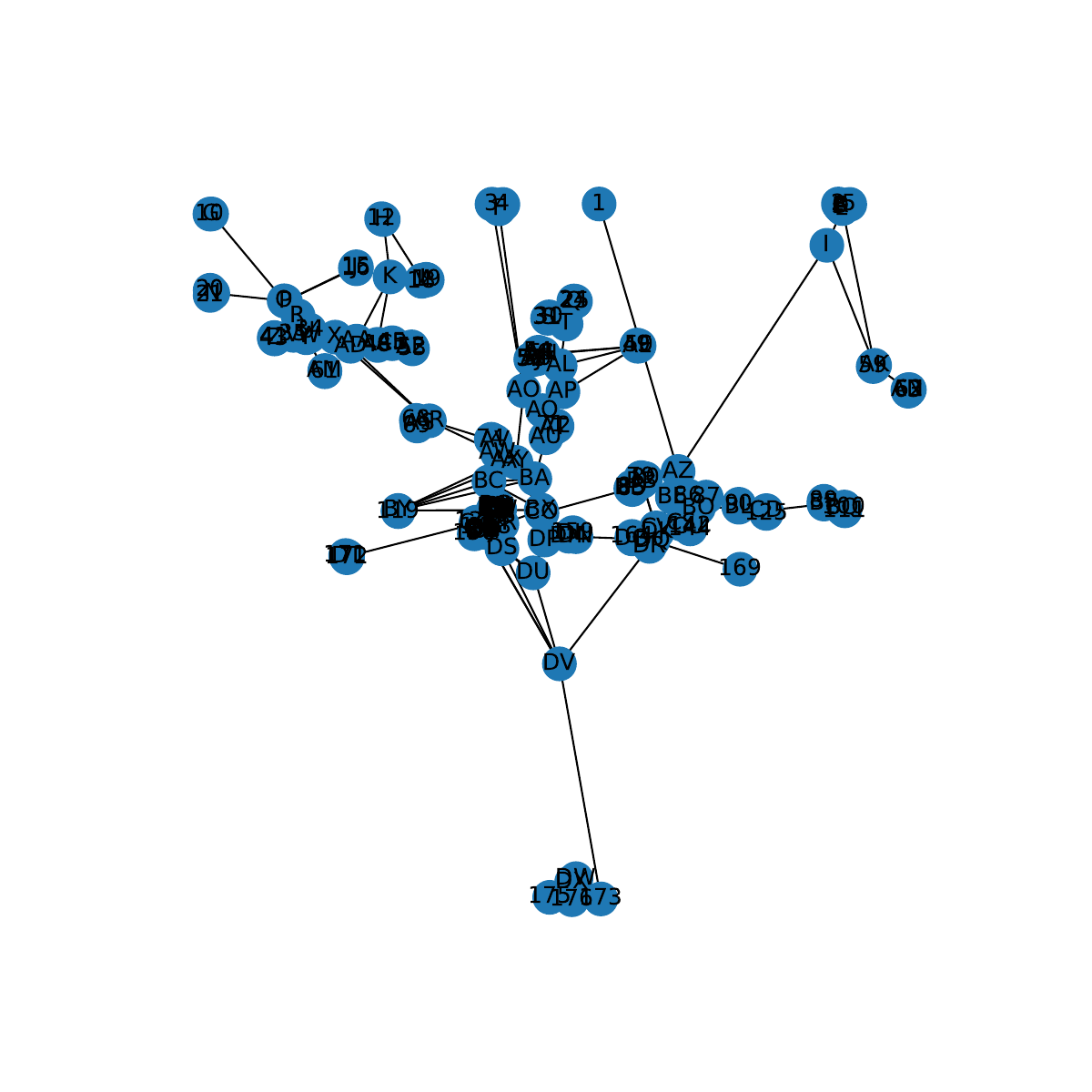}
    \caption{\centering Graph representation of example tree (pre-pruning), created using NetworkX.}
    \label{fig:node_graph}
\end{figure}

By abstracting the tree structure as a network, where branch intersections are represented as nodes and where branches are represented as edges that connect between nodes, the tree structure has been been transformed such that it can be effectively analysed with graph theory. We consider that the application of graph theory is appropriate at this stage of the image processing pipeline because there is no limit on the number of possible edges that connect between nodes and because the edges do not possess any directional quality. The term 'degree' refers to the number of edges that connect at a node.
By identifying branches present in a 4x4 kernel around each intersection point, the connectivity information is formatted into node and edge lists, to create a graph using the open-source package NetworkX \cite{SciPyProceedings_11}. This graph can be visually represented as shown in Figure \ref{fig:node_graph}.

\subsubsection{Calculate Weightings}

In order to 'prune' the structure, a tailored method for aerial drone perching is employed to set thresholds that can be applied to a diverse dataset of tree types. A 'weighting' metric is calculated using the branch lengths and widths that are all normalised to the tree's extrema.

\noindent

The branch weighting is given by:
\begin{equation}
W_b \;=\; \alpha\,\frac{l}{L_{\max}}
       \;+\; (1-\alpha)\,
             \underbrace{\left(\frac{1}{n}\sum_{i=1}^{n}\delta_i\right)}_{\text{in-spec fraction}}
             \,\frac{\overline{w}}{W_{\max}},
\qquad \alpha \in [0,1].
\label{eq:weighting}
\end{equation}
The indicator function $\delta_i$, marking whether the local branch
width lies within the drone-specification thresholds, is defined as:
\begin{equation}
\delta_i \;=\;
\begin{cases}
  1, & w^{\rm spec}_{\min} \le w_i \le w^{\rm spec}_{\max}, \\[2pt]
  0, & \text{otherwise.}
\end{cases}
\label{eq:indicator}
\end{equation}
\noindent\textbf{Variables:}
\begin{itemize}
  \item $\alpha \in [0,1]$ is the length-to-width bias factor, set to
        $0.6$ for the work described.
  \item $l$ is the length of the branch (in pixels).
  \item $L_{\max}$ is the length of the longest branch in the tree
        (in pixels).
  \item $n$ is the total number of pixels along the branch.
  \item $\delta_i$ (Equation~\ref{eq:indicator}) is $1$ if $w_i$ lies
        within the drone-specification range, and $0$ otherwise.
  \item $w_i$ is the branch width at pixel $i$ (in pixels).
  \item $\overline{w}$ is the average branch width (in pixels).
  \item $w^{\rm spec}_{\min},\, w^{\rm spec}_{\max}$ are the lower and
        upper drone-specification width thresholds (in pixels).
  \item $W_{\max}$ is the maximum branch width observed in the tree
        (in pixels).
\end{itemize}

\subsubsection{Prune Graph}

Two main functions are performed during pruning:
\begin{enumerate}
    \item Combine together nodes of degree 2 in series. This acts as a smoothing function to simplify the structure after pruning.
    \item Prune edges below the threshold weighting, given they are 'end point' nodes (where the edge must be connected to a node of degree 1).
\end{enumerate}

Pruning is performed iteratively until there are no changes in the graph from the previous iteration.

\subsection{Profiling and Analysis}\label{step3}
\subsubsection{Analyse Branches}\label{analyse_branches}

In order to profile the possible perching spots along each branch section, a sliding window method is used to iteratively yield branch pixels at a specified width (the approximate width of the drone). The relevant parameters are calculated for every window; the procedure is summarised for each in Table \ref{tab:parameter_methods}:

\begin{table}[h]
    \centering
        \caption{Methods for Parameter Calculation}
    \begin{tabular}{lp{12cm}}
        \hline
        \textbf{Parameter} & \textbf{method} \\
        \svhline
        Angle & Fit a first-order polynomial and return the gradient of the equation. Angle is determined using the equation: 
        \[
        \theta \;=\; \left|\,\operatorname{atan2}\!\bigl(y_n - y_1,\; x_n - x_1\bigr)\,\right| \cdot \frac{180^{\circ}}{\pi}
        \]
        where $(x_1,y_1)$ and $(x_n,y_n)$ are the branch endpoints. This is robust to vertical branches and gives an unsigned angle to the horizontal directly. \\
        \hline
        Curvature & Compute first and second derivatives of the (\textit{x}, \textit{y}) coordinates. Curvature is given by: 
        \[
        \text{curvature} = \kappa(t) \;=\; \frac{\dot x(t)\,\ddot y(t) \;-\; \dot y(t)\,\ddot x(t)}
                     {\bigl(\dot x(t)^2 + \dot y(t)^2\bigr)^{3/2}}
        \]
        where $t$ indexes pixels along the branch in their ordered traversal, and $\dot{}$ denotes the central finite difference. A moving average filter is used to smooth the curvature values, returning an array of curvature values along the line. \\
        \hline
        Widths & Return the minimum, maximum, and average width along the branch. \\
        \hline
    \end{tabular}
    \label{tab:parameter_methods}
\end{table}

Thresholds for each parameter are used to filter the sections into a list of viable options. A curvature threshold was experimentally determined by probing the result of the curvature calculation for a variety of trees in the dataset. A high curvature value indicates not only that the branch could have some imperfections in its shape, but also that the angle calculation (which assumes a linear branch shape) will likely be unrepresentative. The thresholds are designed to be lenient to guarantee that, within reason, at least a handful of location options are selected for a variety of trees.

\subsubsection{Rank Viable Branches}

In order to determine a single ideal landing spot, the PLI method ranks the remaining viable branches according to their suitability. Because the criteria used to rank the branches are independent of one another, the Multi-Criteria Decision Making (MCDM) method was chosen as a penalty function. This method takes each parameter (angle, curvature, average width) normalised by the range of the parameter thresholds, and then applies a 'penalty' based on how far this value is from the 'perfect' branch. The perfect branch is represented as one with an angle of 0 $\degree$, 0 curvature, and an average width at the lower bound of the width threshold. \textsuperscript{1} \footnote{\textsuperscript{1} Experimental findings from Li's paper \cite{li_tendon-driven_2025} show that while the claw can grasp a branch radius range of 30 to 110 mm, there is a negative linear relationship between the width of the branch and the possible payload.}
Concretely, the penalty assigned to a viable candidate branch $b$ is
the weighted sum of normalised absolute deviations from the ideal:
\begin{equation}
P(b) \;=\; \sum_{j \,\in\, \{\text{angle},\,\text{width}\}}
           \lambda_j\,
           \frac{\bigl|\,p_j(b) - p_j^{\star}\,\bigr|}
                {p_j^{\max} - p_j^{\min}},
\qquad
\sum_{j} \lambda_j = 1, \quad \lambda_j \ge 0.
\label{eq:mcdm}
\end{equation}
The branch minimising $P(b)$ over the set of viable candidates is
selected as the perching location.

\noindent\textbf{Variables:}
\begin{itemize}
  \item $b$ is a candidate branch segment from the set of viable
        branches.
  \item $p_j(b)$ is the value of criterion $j$ measured on branch $b$:
        the absolute angle $|\theta(b)|$ for $j=\text{angle}$
        (Table~\ref{tab:parameter_methods}), and the average width
        $\overline{w}(b)$ for $j=\text{width}$.
  \item $p_j^{\star}$ is the ideal value of criterion $j$, with
        $p^{\star}_{\text{angle}} = 0^{\circ}$ and
        $p^{\star}_{\text{width}} = w^{\rm spec}_{\min}$. The minimum
        in-spec width is preferred to maximise the available payload
        margin, as discussed below.
  \item $p_j^{\min},\, p_j^{\max}$ are the lower and upper viability
        thresholds for criterion $j$, so the denominator
        $p_j^{\max} - p_j^{\min}$ is the width of the acceptable range
        and normalises each criterion to the unit interval.
  \item $\lambda_j \in [0,1]$ is the bias weight for criterion $j$,
        constrained so that $\sum_j \lambda_j = 1$ and consequently
        $P(b) \in [0,1]$.
\end{itemize}
The curvature criterion is retained as a viability filter in
Section~\ref{analyse_branches} but is excluded from
Equation~\eqref{eq:mcdm} for the reasons discussed in
Section~\ref{exemplar_results}.
Consequently, thinner branches have been prioritised here to allow for a larger payload, given that branches which are below the threshold have already been filtered out. The ranking bias for each parameter can be adjusted by specifying
the coefficient values $\lambda_{\rm angle}$ and $\lambda_{\rm width}$,
subject to $\lambda_{\rm angle} + \lambda_{\rm width} = 1$. For example, the effect of increasing the bias towards branch angle (value of $\lambda_{\text{angle}}$) will result in branches with angles close to 0$\degree$ being ranked preferentially over options with equally ideal widths or curvature.
The final output is in the format of a pixel coordinate at the mid-point of the selected segment.

\subsubsection{Branch Diameter and Structural Integrity}

As described previously, the threshold for minimum branch diameter required for the PLI method is given as 30 mm. As determined from the work of Li et al., 30 mm represents the branch diameter at which the clamping force generated by the tendon-driven claw is no longer sufficient for perching to occur. \cite{li_tendon-driven_2025}
In addition to affecting the clamping forces that can be generated, branch diameter also plays a substantive role in determining the bending stress that a tree branch undergoes during loading. Consequently, the PLI method must ensure that the
specification threshold $w^{\rm spec}_{\min}$ is significantly
greater than would be required to support the perching drone.

\begin{equation}
    \sigma_{\max} \;=\; \frac{M\,r}{I},
\qquad
M \;=\; F\,L,
\qquad
I \;=\; \frac{\pi r^{4}}{4}\label{eq:moment_inertia}
\end{equation}



                        
Variables:
\begin{multicols}{2}
    \begin{itemize}
        \item $M$ is the moment generated by the object on the branch. 
        \item \( F \) is the perpendicular force at perching location.
        \item \( L \) is the horizontal distance from the branch origin.
        \item \( r \) is the radius of the branch.
        \item \( I \)  is the moment of inertia for a cylindrical beam.
        \item \( \sigma \) is the bending stress produced.
    \end{itemize}
\end{multicols}

\begin{center}
 
\end{center}

From equation \eqref{eq:moment_inertia}, we can calculate that the bending stress that results from perching a drone weighing 1.5 kg on a 15 mm radius branch, 2 m from the base of the branch, will be approximately 11.1 MPa. This is significantly lower than the reported bending stress required for such a branch to fail through rupture. According to the Wood Handbook survey of the mechanical properties of 88 North American trees, only two had a Modulus of Rupture (MoR) for bending of less than 30 MPa (29 MPa and 27 MPa) \cite{WoodHandbook}. Consequently, to reduce computational overhead, we have omitted a branch bending stress check for the current implementation of the PLI method.

\section{Experimental Validation}

Although our methodology has focussed on the performance of our PLI method pipeline rather than field experimentation, we were able to demonstrate the real-world potential of the PLI method by acquiring images of trees in a forest environment, using the PLI method to determine the optimal branch and then manually perching the drone on the branches designated as optimal.

\begin{figure}[ht]
    \centering
    \begin{subfigure}[b]{0.32\textwidth}
        \centering
        \includegraphics[width=\textwidth]{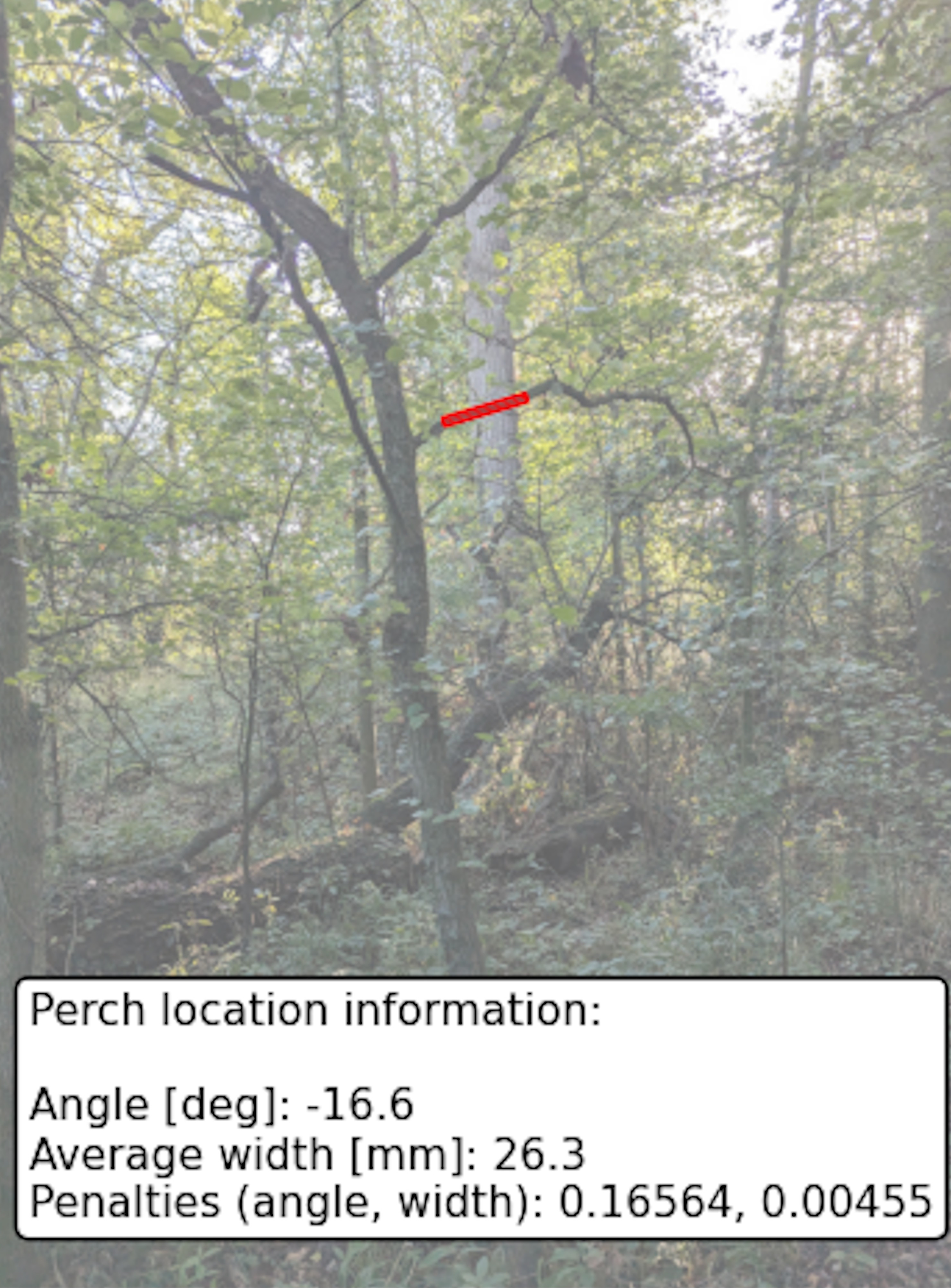}
        \caption{First forest tree analysed with PLI.}
        \label{fig:subfig1}
    \end{subfigure}
    \hfill
    \begin{subfigure}[b]{0.32\textwidth}
        \centering
        \includegraphics[width=\textwidth]{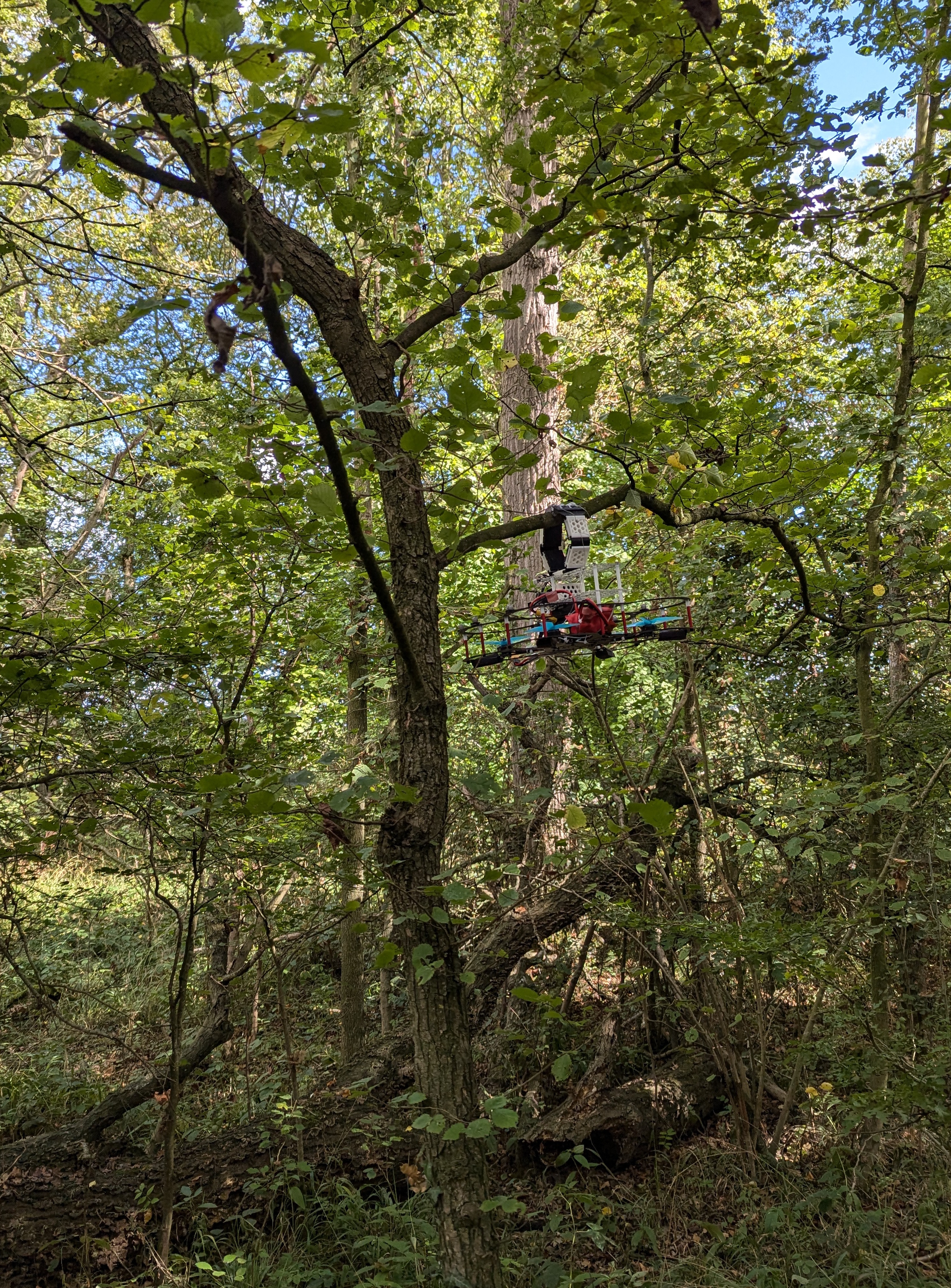}
        \caption{Tendon-driven drone static perching on first forest tree.}
        \label{fig:subfig2}
    \end{subfigure}
    \hfill
    \begin{subfigure}[b]{0.32\textwidth}
        \centering
        \includegraphics[width=\textwidth]{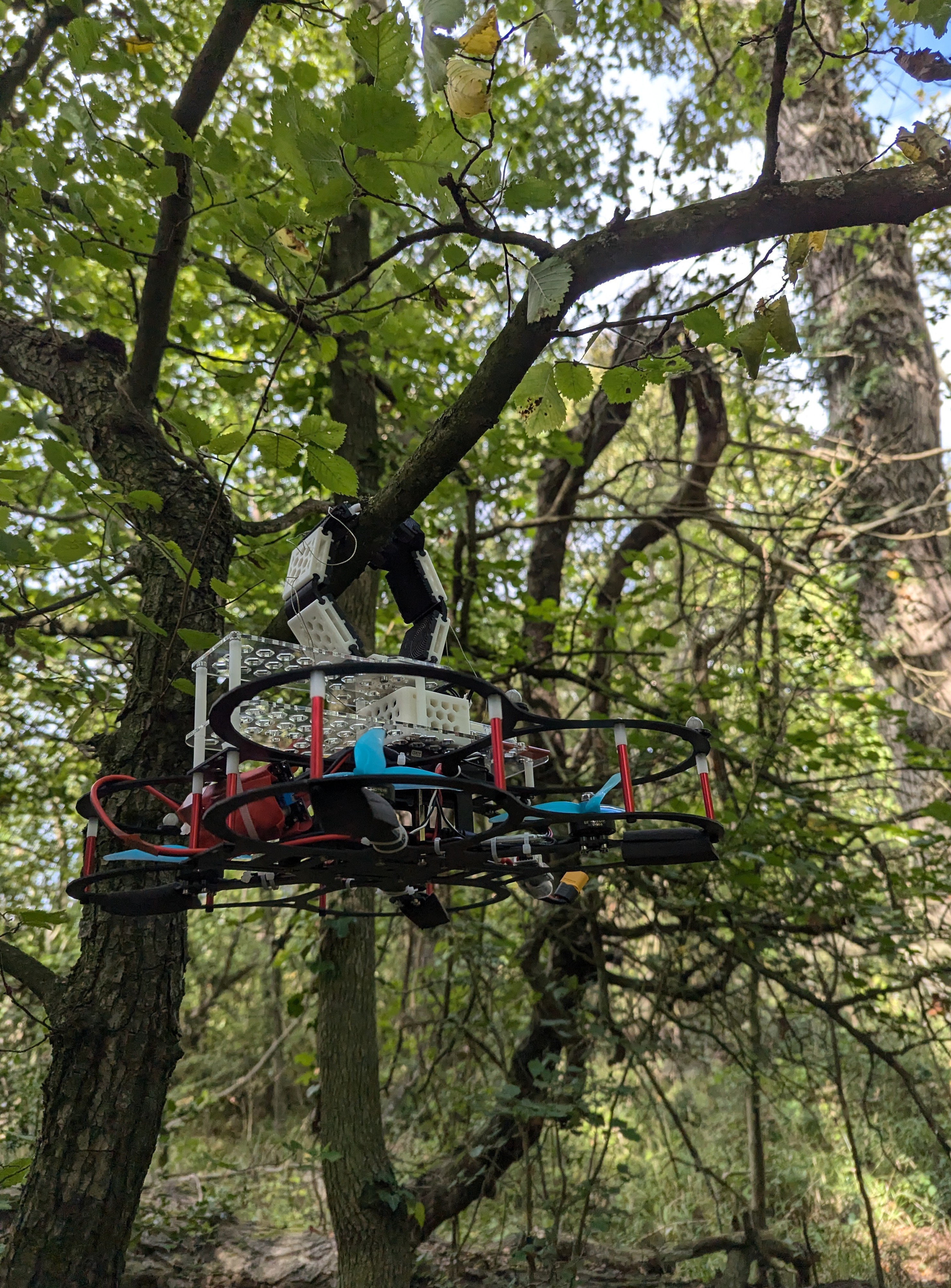}
        \caption{Close-up of drone perching on first forest tree.}
        \label{fig:subfig3}
    \end{subfigure}
    
    \vspace{0.5cm} 
    
    \begin{subfigure}[b]{0.32\textwidth}
        \centering
        \includegraphics[width=\textwidth]{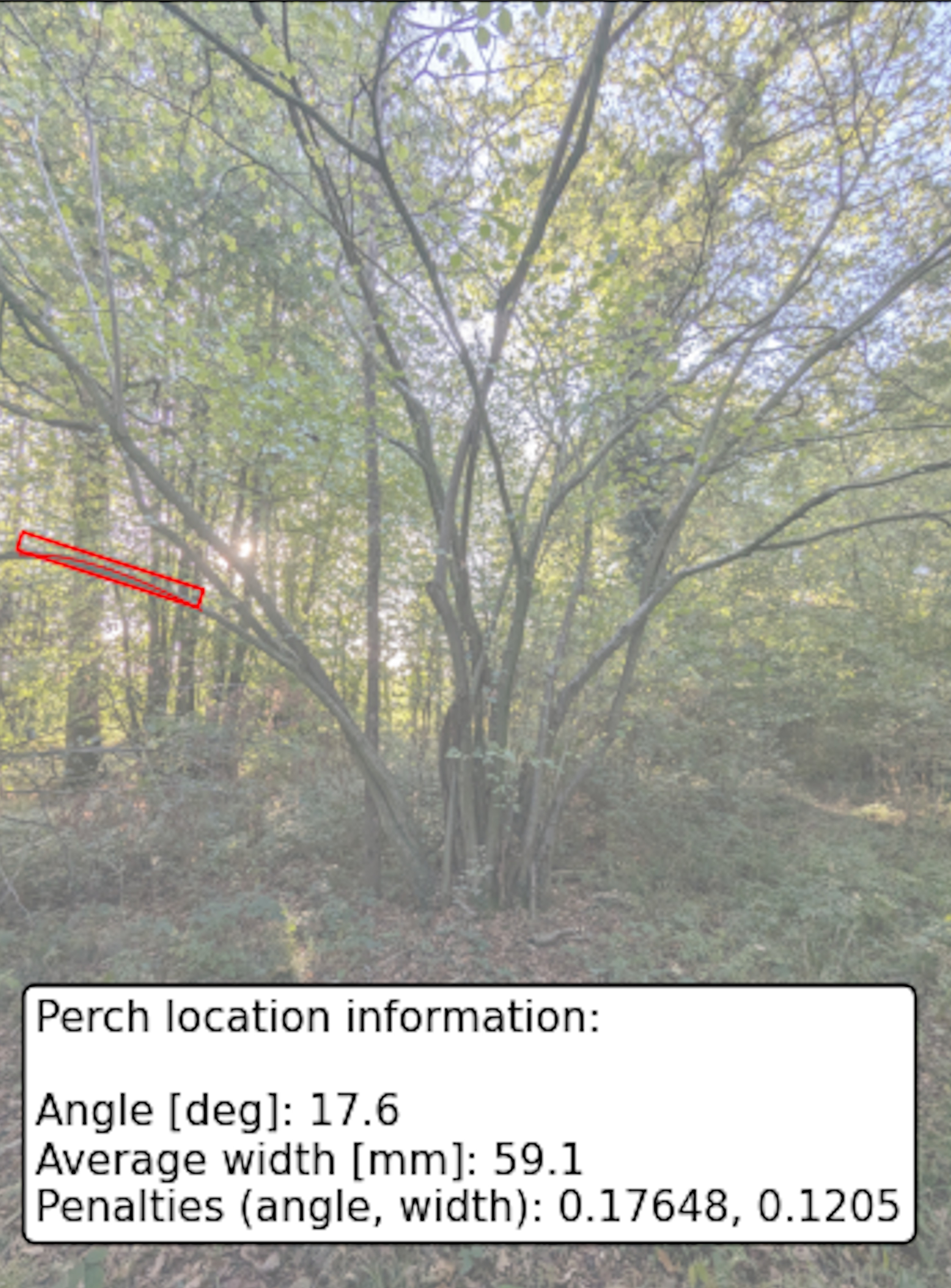}
        \caption{First forest tree analysed with PLI.}
        \label{fig:subfig4}
    \end{subfigure}
    \hfill
    \begin{subfigure}[b]{0.32\textwidth}
        \centering
        \includegraphics[width=\textwidth]{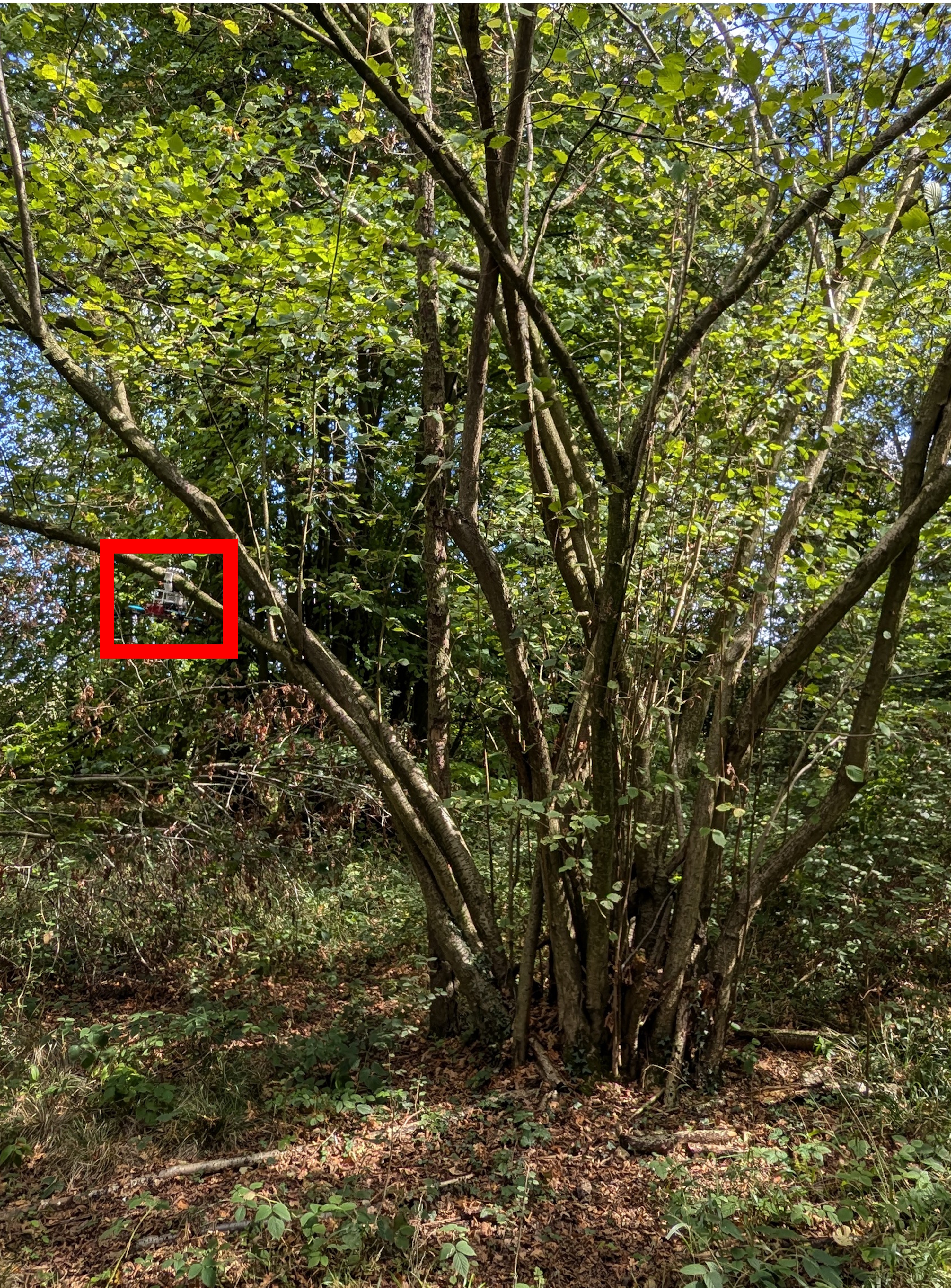}
        \caption{Tendon-driven drone static perching on second forest tree.}
        \label{fig:subfig5}
    \end{subfigure}
    \hfill
    \begin{subfigure}[b]{0.32\textwidth}
        \centering
        \includegraphics[width=\textwidth]{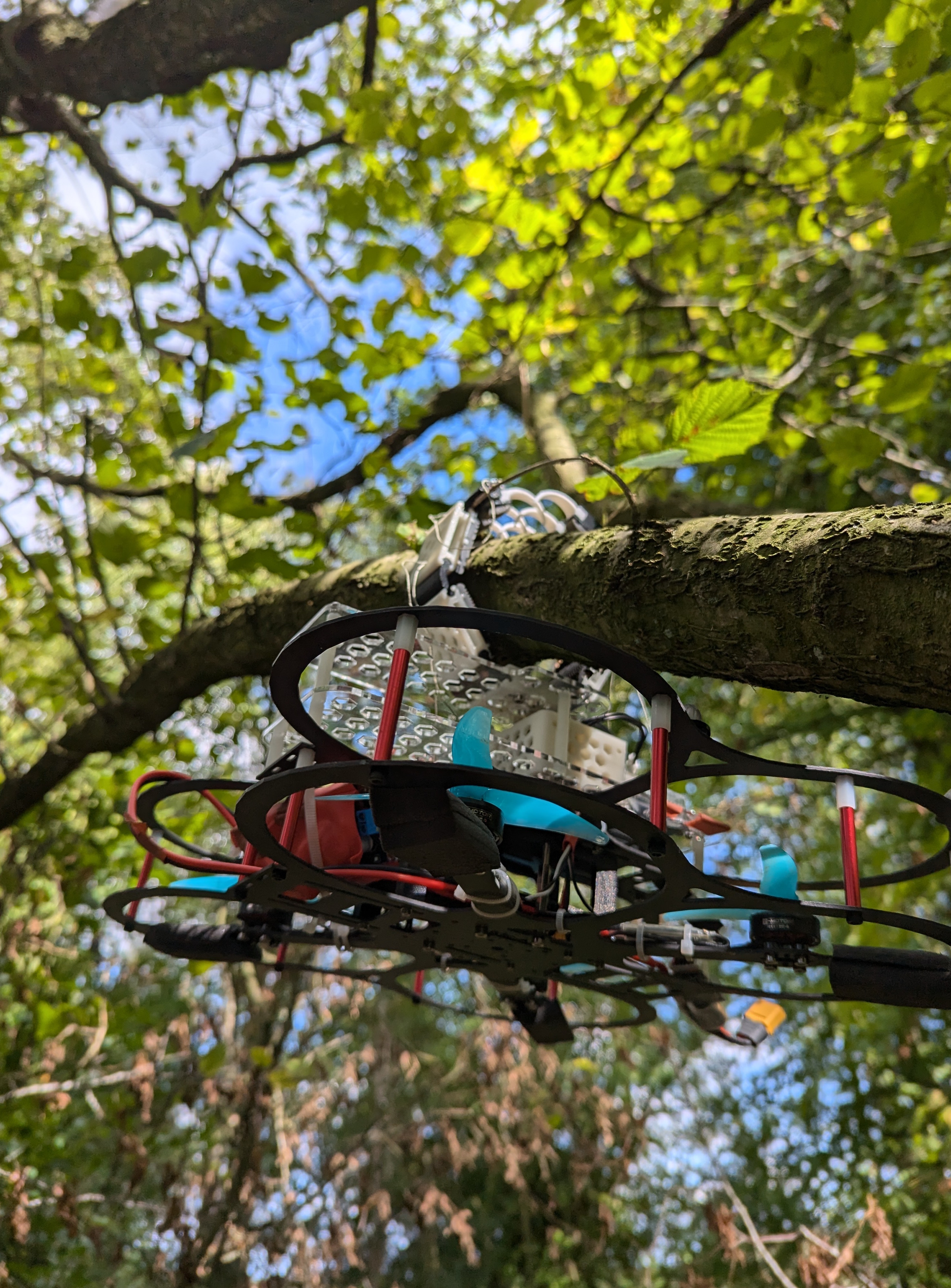}
        \caption{Close-up of drone perching on second forest tree.}
        \label{fig:subfig6}
    \end{subfigure}
    
    \caption{Demonstrating manual perching of tendon-driven grasping drone on locations informed by PLI method.}
    \label{fig:perchinginpaulerspury}
\end{figure}

To acquire the images of trees in the forest environment, a Google Pixel 7 image sensor was used. The images were given as inputs to the PLI method, running onboard a Dell Latitude 5450 Intel Core Ultra 7 laptop and the output images can be seen in Figure \ref{fig:subfig1} and Figure\ref{fig:subfig2}. As a solution for robust branch segmentation for PLI method has not yet been developed, the segmentation masks for these newly acquired images were illustrated by hand. Optimal perching locations were produced as the output. To test the tendon-driven grasping drone's ability to statically perch on the selected branches, the drone was manually moved into position underneath the branch and the tendon-driven claw was manually retracted and released onto the relevant branch. The static perch was deemed successful if the drone remained perched for 10 seconds after the initial grasp. As shown in Figure \ref{fig:subfig2} and Figure \ref{fig:subfig4}, we were able to determine that, at least in two instances, the optimal perching locations recommended by the algorithm were sufficient for static perching with a tendon-driven grasping drone.

The field validation test was performed at the start of autumn at co-ordinates 52.0871°N, 0.9450°W, in a temperate woodland located on Plumpton End Road, Paulerspury, Northamptonshire, UK.

\section{Results}

\begin{figure}[H]
    \centering
    \includegraphics[width=\textwidth]{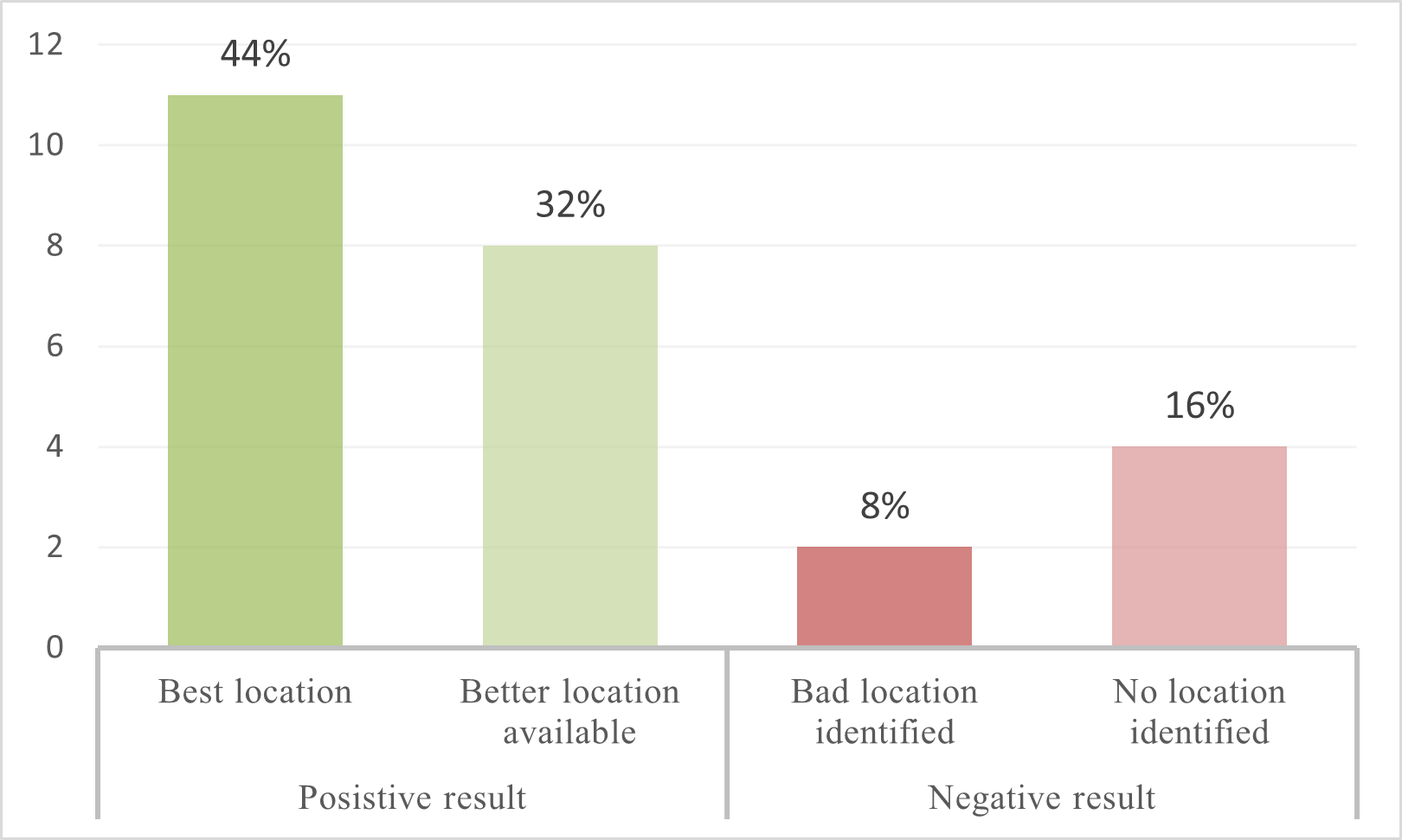}
    \caption{Results for a validation sample set, categorised into 4 possible outcomes.}
    \label{fig:result-bar}
\end{figure}

As can be seen from Figure \ref{fig:result-bar}, the PLI method achieved success in 76\% of the instances where the 22 dataset images that comprise the validation set were given as inputs. In the case of a failure to identify a perching location, an exploration of failure modes is given in Table \ref{tab:incorrect-table}.

\clearpage
\subsection{Incorrect Result Examination}
\begin{table}[H]
    \centering
        \caption{Three examples of incorrect or non-viable perching locations.}
    \begin{tabular}{|c||p{6cm}|p{4cm}|}
        \hline
        \textbf{Erroneous Example Image} & \textbf{Reason for issue} & \textbf{Possible Fix} \\
        \hline \hline
        \begin{minipage}[t]{0.29\textwidth}
            \vspace{0pt}
            \includegraphics[trim=18cm 2.6cm 16.5cm 2.7cm, clip, width=\linewidth]{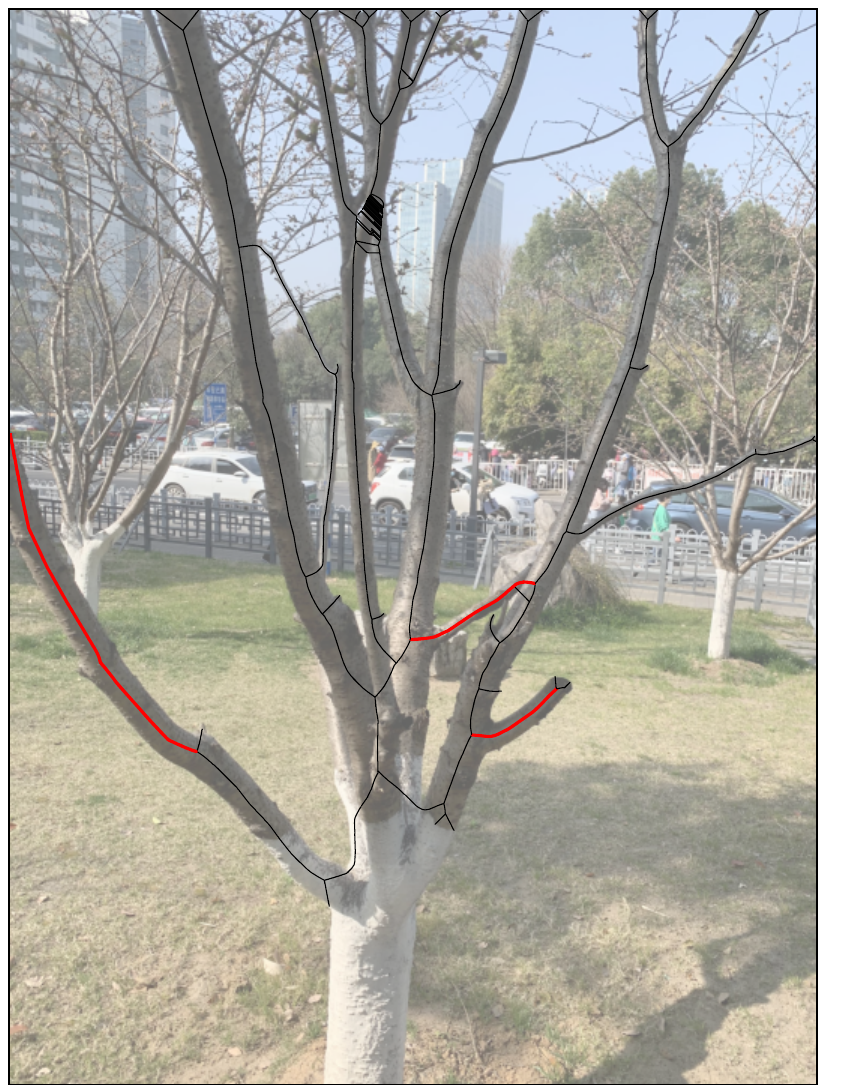}
            \vspace{0pt}
        \end{minipage} 
        & \begin{minipage}[t]{6cm}
            \raggedright
            \vspace{0pt} 
            This plot has highlighted entire branches which contain viable regions. 
            
            The skeleton for this example is incorrect because the dataset does not include the white region of the trunk in the segmentation mask. In this particular example, the PLI method is still able to identify some possible options for perching location, but this would otherwise cause issues with the specific algorithm used to acquire the pixel to length ratio, thus making all width values incorrect.
        \end{minipage} 
        & \begin{minipage}[t]{4cm}
            \vspace{0pt} 
            \raggedright
            Amend the dataset so that the segmented mask includes regions that are covered in white paint.
        \end{minipage} \\
        \hline
        \begin{minipage}[t]{0.29\textwidth}
            \vspace{0pt}
            \includegraphics[trim=18cm 2.6cm 16.5cm 2.7cm, clip, width=\linewidth]{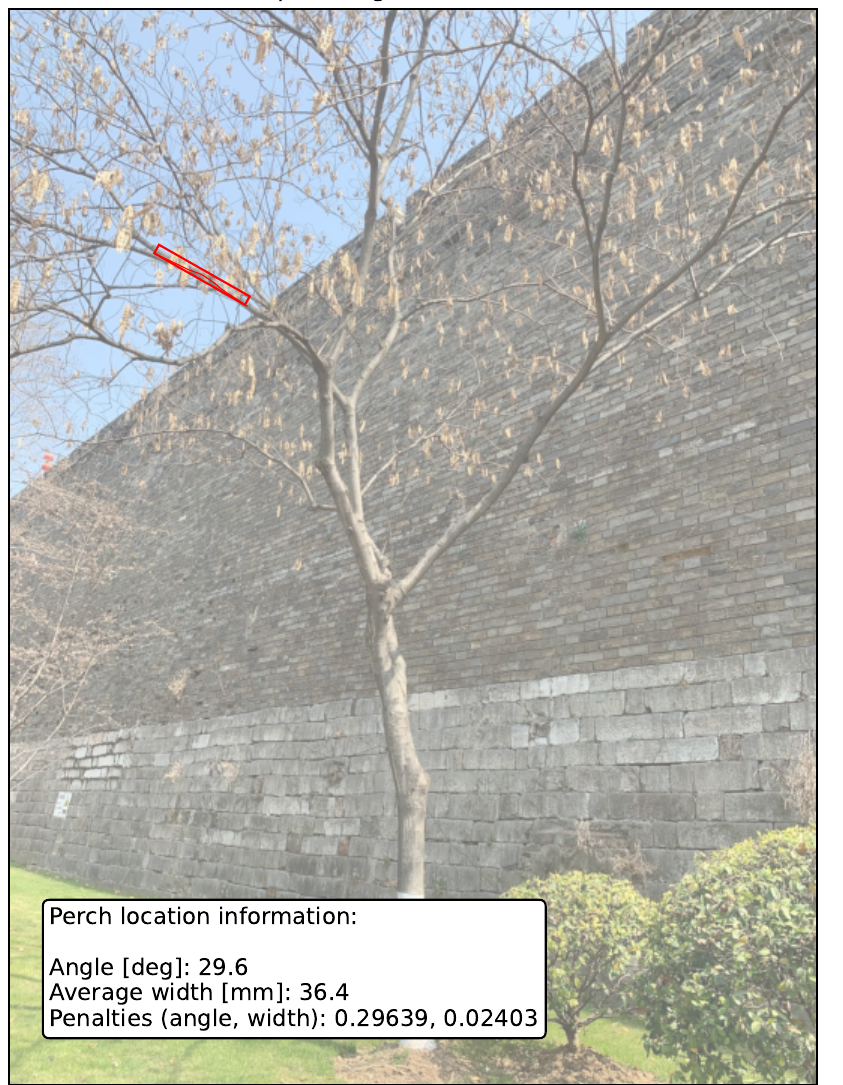}
            \vspace{0pt}
        \end{minipage}  
        & \begin{minipage}[t]{6cm}
            \raggedright
            \vspace{0pt}
            This plot shows the perch location result from the method (with $\lambda_{\text{angle}}$ = 0.8).
            This result is valid for the drone's specification derived thresholds.
            
            However, in reality this region is densely obstructed with smaller branches and foliage, and would not be an ideal perch location. This is because the process used has no method of identifying this type of obstruction.
        \end{minipage}  
        & \begin{minipage}[t]{4cm}
        \raggedright
            \vspace{0pt}
            This particular edge case is detailed in Section \ref{future-dev}. 
            This dataset primarily features trees with limited foliage, however if there are a significant amount of leaves which fully obscure the underlying branch structure, these regions are omitted from the segmentation mask of the tree. Thus, they are not included in consideration for a perching location from the first step of the process. 
        \end{minipage} \\
        \hline
        \begin{minipage}[t]{0.29\textwidth}
            \vspace{0pt}
            \includegraphics[trim=18cm 2.6cm 16.5cm 2.7cm, clip, width=\linewidth]{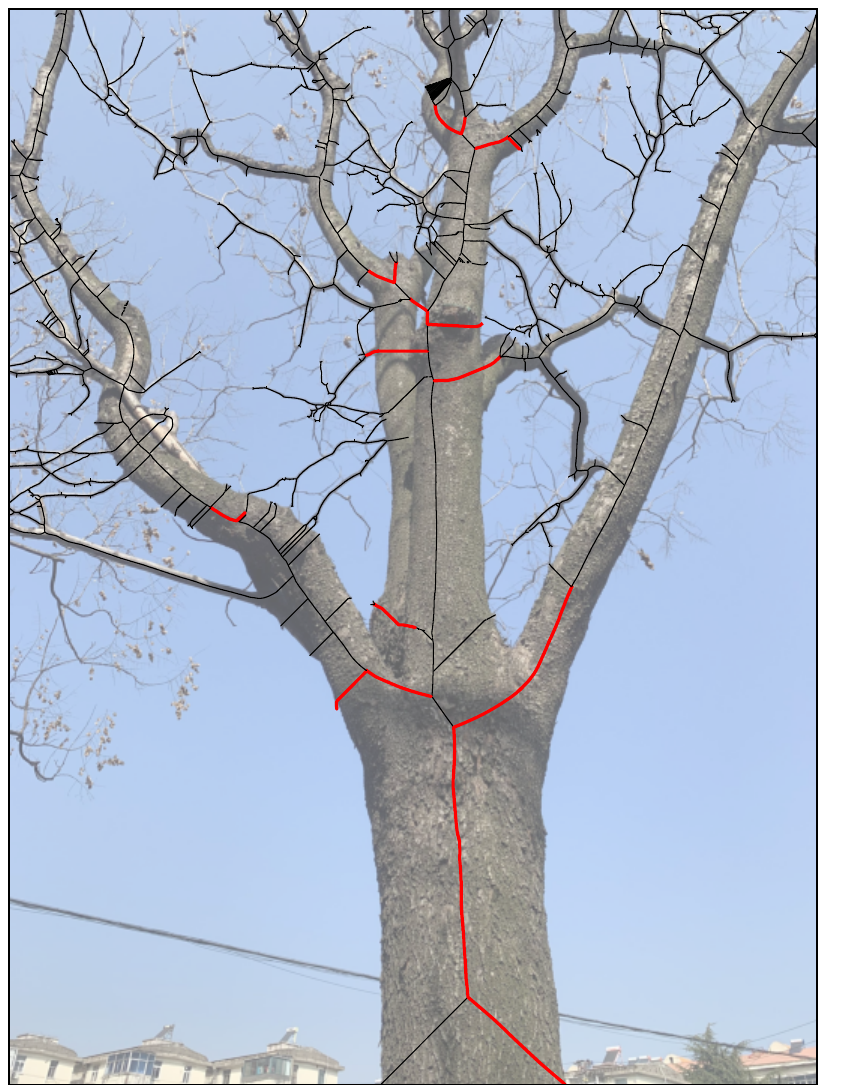}
            \vspace{0pt}
        \end{minipage}  
        & \begin{minipage}[t]{6cm}
            \vspace{0pt}
            This plot shows the computed skeleton and highlights branches that contain viable regions. 
            
            This example is not ideal in a number of ways. For example, sub-optimal image angle can be seen to cause perspective distortion and the segmentation mask has included many minor negligible branches. The skeletonisation algorithm has also produced a number of 'spurious skeleton branches' \cite{saha_chapter_2017} which are derived from boundary noise along the segmentation mask edge.
        \end{minipage}  
        & \begin{minipage}[t]{4cm}
            \vspace{0pt}
            Though these erroneous branches should be removed during pruning, in this case removal may have been insufficient due to an unrepresentative pixel to length ratio being calculated (as a result of the perspective distortion). Alternative methods to effectively smooth the skeleton include 'zig-zag straightening' and 'fusion of close branches' \cite{sanniti_di_baja_chapter_2017}.
        \end{minipage} \\
        \hline
    \end{tabular}
    \label{tab:incorrect-table}
\end{table}

\subsection{Exemplar Results and Method Adjustments}\label{exemplar_results}

A handful of unexpected results occurred during testing, requiring small changes in the analysis and ranking stages: 
\begin{enumerate}
    \item \textbf{Branch reversal:} Not all branches can be considered monotonic, as branch curvature is unpredictable. This would not always be filtered by the curvature parameter due to the smoothing function.
    \noindent As a solution, an additional filter is added which calculates the local gradients along the branch and detects a change in sign, indicating a non-monotonic shape.
    \item \textbf{Curvature Criterion:} Although effective for filtering out branches with extreme contours at the viability check stage, the curvature criterion was found to improperly affect the ranking of viable perch locations. The penalty function MCDM is better suited to scalar parameters, whereas the curvature of the branch is better represented continuously and the average curvature does not give a meaningful description of the branch shape. 
    \noindent As a result, this criteria was removed from the penalty function.
    \item \textbf{Width region:} A number of test cases returned no viable options; generally it is preferred that a perch location is identified that is suboptimal rather than no result at all. \textsuperscript{3} \footnote{\textsuperscript{3} This is the reasoning for splitting the analysis into a viability check and then ranking; some trees which are entirely unsuitable will return no viable options correctly, however the viability criteria should not be too strict due to the inherent uncertainty in this process.}
    \noindent In response to this, after identifying the width criterion as the more frequently failed threshold, a more lenient and logically representative alternative was used. The average width of the branch was calculated over a central region of the section that corresponded roughly to the width of the drone's perching claw. 
\end{enumerate}

The penalty biases (values of $\lambda$ for each criterion) were iteratively adjusted and the results were compared. A clear example of the impact of altering these coefficients is shown in Appendix \ref{app:imgs}, Figure \ref{fig:compare-biases}; A higher $\lambda_{\text{angle}}$ (Figure \ref{fig:high-angle-bias}) results in the selection of a wider and lower slope region, whereas a higher $\lambda_{\text{width}}$ results in the reverse. In context, it is logical to select the higher angle bias as the perching mechanism has a much larger tolerance for a range of branch widths, given that the extreme widths are filtered at the viability threshold stage.

\subsection{Code Performance Investigation}

\begin{figure}[H]
    \centering
    \includegraphics[width=1\linewidth]{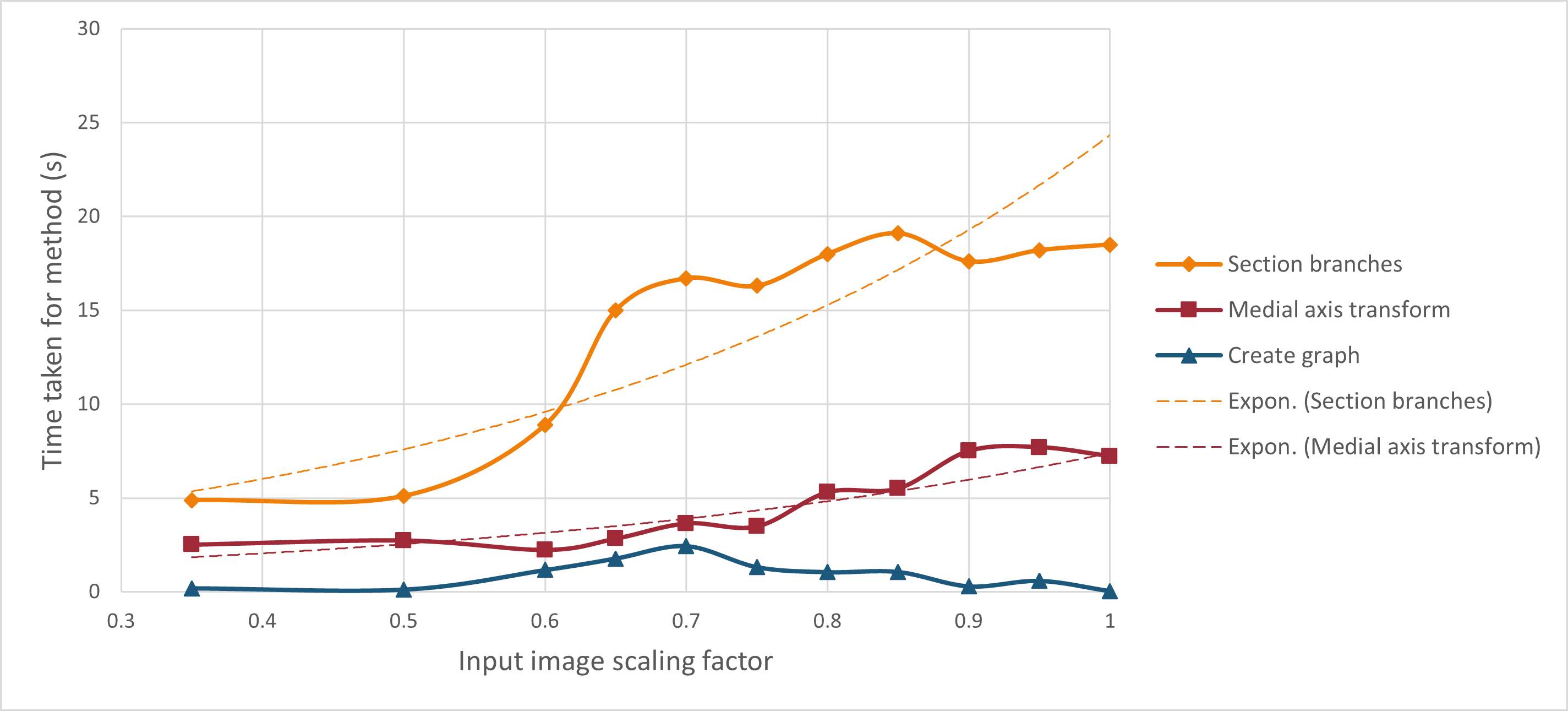}
    \caption{\centering Processing time per method against input image scaling factor. Data obtained using cProfile. \cite{czotter_python_nodate} Performed on Intel Core i7-8665U CPU (4 cores, 8 threads, 1.90GHz) with 16GB RAM.}
    \label{fig:code-prof}
\end{figure}

\begin{figure}[H]
    \centering
    \begin{subfigure}{0.9\textwidth}
        \centering
        \includegraphics[trim=4cm 7cm 1cm 3cm, clip, width=\linewidth]{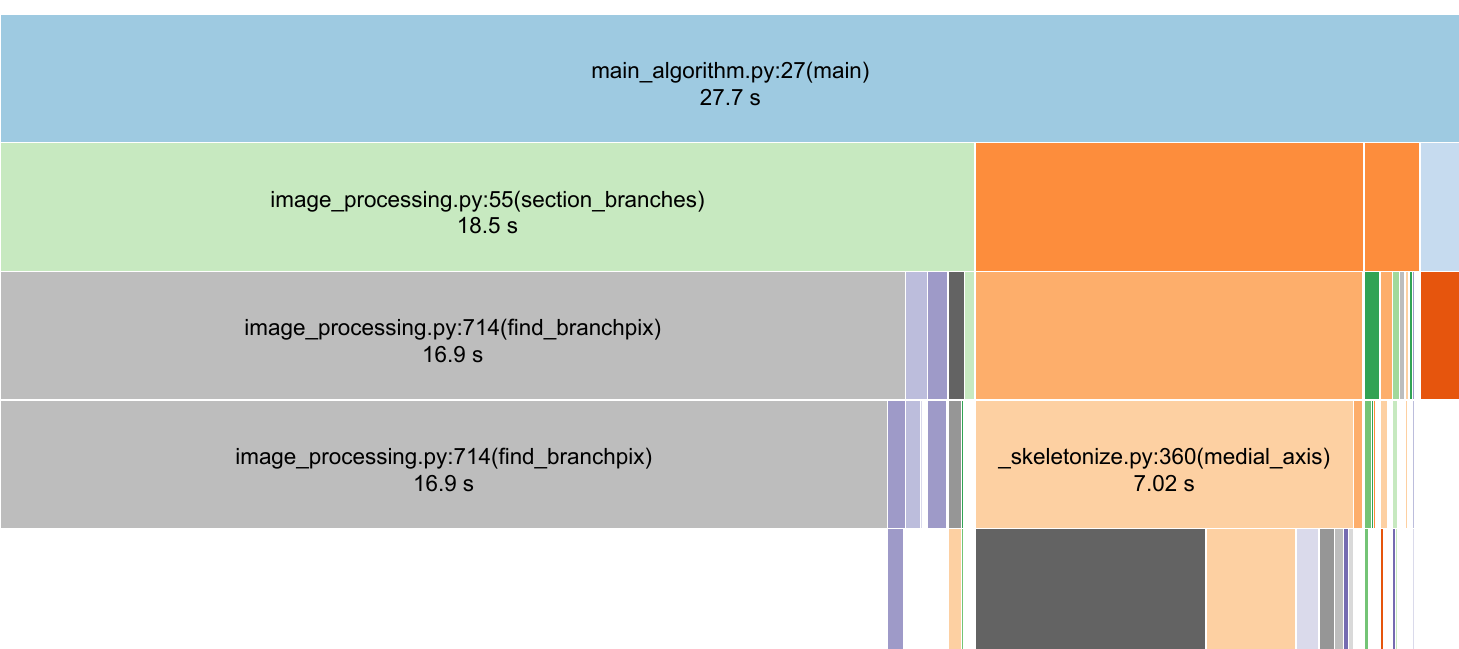}
        \caption{Excludes graph pruning}
        \label{fig:icicle-noprune}
    \end{subfigure}
    \begin{subfigure}{0.9\textwidth}
        \centering
        \includegraphics[trim=7cm 9.5cm 4.5cm 3cm, clip, width=\linewidth]{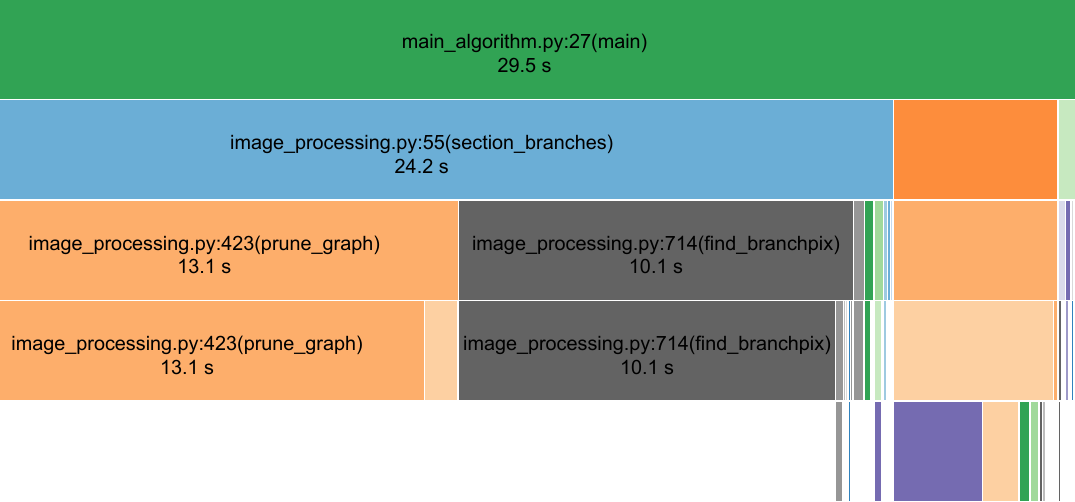}
        \caption{Includes graph pruning}
        \label{fig:icicle-prune}
    \end{subfigure}
    \caption{\centering Snakeviz full code cProfile visualisations. Each layer represents the nested hierarchy of the scripts and methods. Both inputs were at full scale resolution.}
    \label{fig:SV-vis}
\end{figure}

Utilising the data provided by cProfile, \cite{czotter_python_nodate} a built-in Python package for surveying the processing time for each algorithm in the PLI method, it is possible to obtain a holistic view of where inefficiencies are located. The package 'Snakeviz' \cite{bray_snakeviz_2012} is used to produce an 'icicle' representation of each function and sub-function (see Figure \ref{fig:code-prof}). Some key characteristics were identified: 
\begin{enumerate}
   \item \textbf{Input scaling:} The code allows for an input parameter that specifies a scaling factor which can reduce the resolution of the input image for efficiency. Excluding the graph pruning method, the next two most time-costly methods are the 'find\_branchpix' (as part of the parent 'section\_branches' method) and the medial axis transformation. This result is constructive as both of these methods are fundamentally pixel morphology techniques, where a pixel kernel scans the whole image. Hence, it is logical that the processing time would be directly reduced with the resolution of the input image. Figure \ref{fig:code-prof} supports this hypothesis. Note that there is a large variance with each run of the code and that the exact processing times are dependent on the system's hardware, so this chart serves to show the trend rather than the absolute processing times.
    \item \textbf{Graph pruning:} As seen in Figure \ref{fig:SV-vis} the graph pruning stage consistently occupied the largest portion of processing time.  This result called into question the necessity of this stage. The code was modified to bypass this method, and the results were compared. The code without pruning included is shown in Figure \ref{fig:icicle-noprune} and the code with graph pruning included is shown in Figure \ref{fig:icicle-prune}. For all tested examples, there was no example of the result significantly changing as a result of not pruning the graph. However, at this stage, the branch connectivity information is not utilised at the branch analysis stage; if this information is required for any future developments, this pruning stage is vital in order to create a graph that is representative of the tree. 
\end{enumerate}

\section{Discussions}

\subsection{Method Limitations and Issues}

\subsubsection{Image Segmentation}
\begin{figure}[ht]
     \centering
     \begin{subfigure}[b]{0.45\textwidth}
         \centering
         \includegraphics[width=\textwidth]{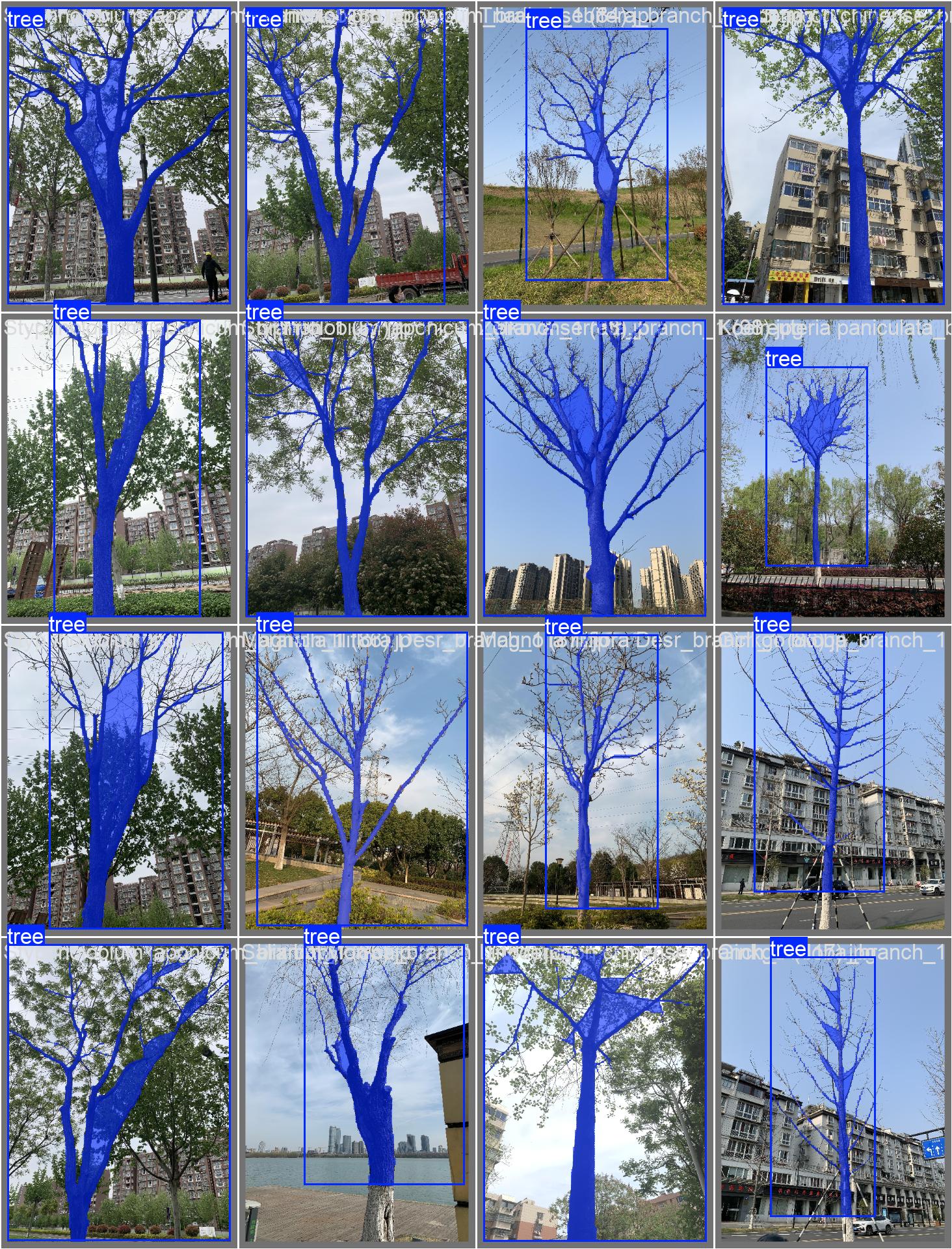}
         \caption{Dataset Labels (Validation Input)}
         \label{fig:val_labels}
     \end{subfigure}
     \begin{subfigure}[b]{0.45\textwidth}
         \centering
         \includegraphics[width=\textwidth]{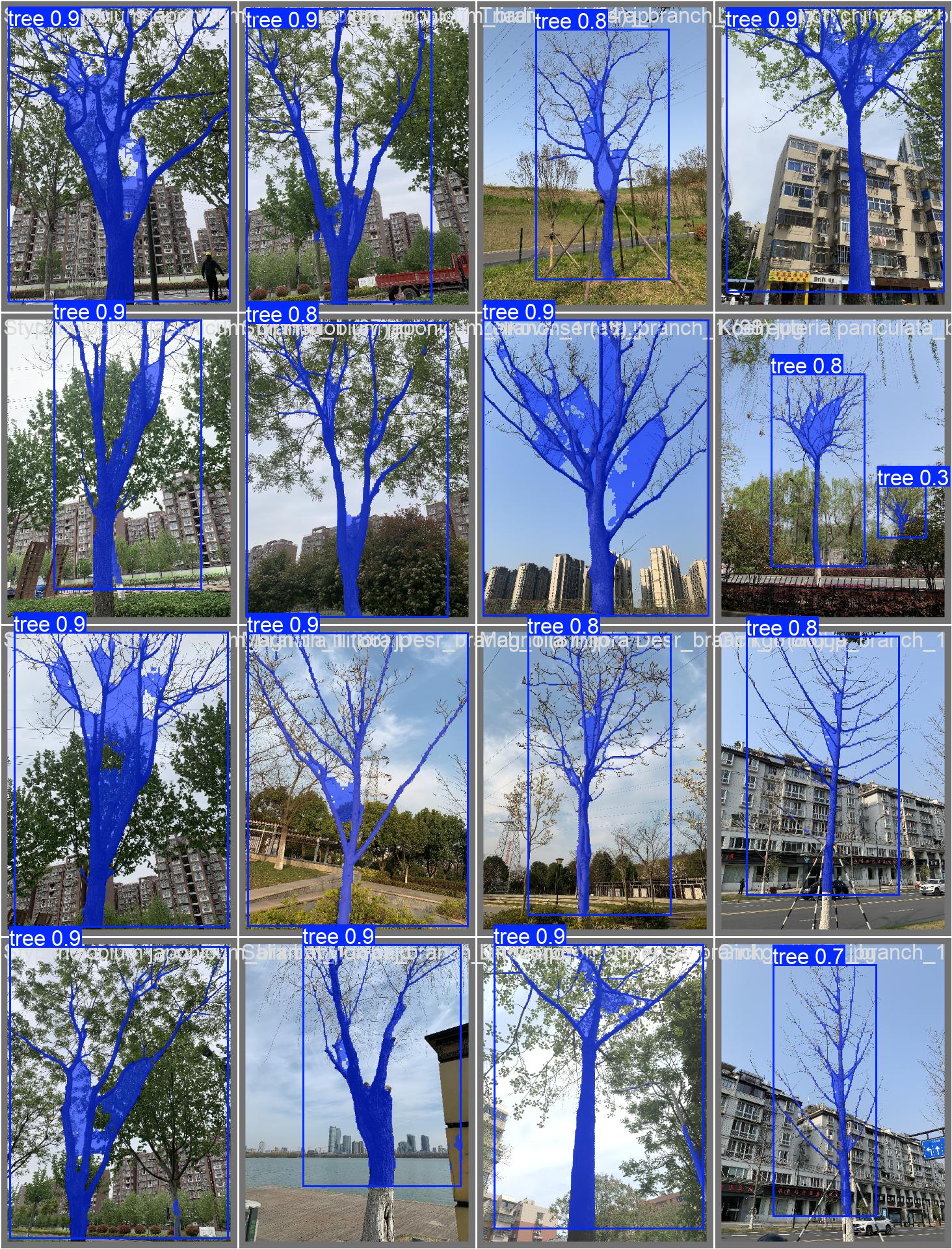}
         \caption{Validation model outputs}
         \label{fig:val_pred}
     \end{subfigure}
        \caption{\centering YOLOv11 Training results. Fig. \ref{fig:val_labels}: the input data, designated validation, as interpreted by the model. Fig. \ref{fig:val_pred}: the output of the model after 20 training epochs.}
        \label{fig:training_out}
\end{figure}

A significant portion of the study was dedicated to preparing the dataset and training the model, as well as troubleshooting errors that occurred during the latter. A critical issue was the model's interpretation of the dataset input masks; in areas where branches overlapped creating enclosed loops in the mask, the model could not accommodate the enclosed negative space. Despite creating a workaround method (see Appendix \ref{convert-dataset}), the model continued to interpret it incorrectly. Figure \ref{fig:val_pred} shows a set of results for the provided input masks (Figure \ref{fig:val_labels}), with the enclosed regions incorrectly included as part of the mask. 

Given the scope of this study, the decision was made to continue development using dataset masks (for example, see Figure \ref{fig:mask}) to simulate the output of a correct segmentation model. Previous research indicates that machine learning can perform similar segmentation tasks \cite{li_tree_2023}, but in this case, we found that Ultralytics YOLOv11 was not viable. Future efforts should use models specifically designed for segmentation, such as DeepLabV3 \cite{pytorch_deeplabv3_nodate} or U-Net. \textsuperscript{2} \footnote{\textsuperscript{2} Ultralytics specialises in object detection and tracking, whereas DeepLabV3 and U-Net are more suited for segmentation.}

\subsubsection{Foliage Obscuration}

A limitation of the proposed PLI method is the requirement that branches be visible such that available branches can be accurately segmented. In some cases, the presence of foliage can obscure branches in an acquired image and prevent accurate segmentation. This is a challenge that has been explored by a variety of authors in the field. Chen et al. explored the challenges of fruit tree pruning and fruit thinning where the fruit were occluded by leaves. Most recently, Geckeler et al. explored how to learn the depth of branches that are otherwise excluded, for the purpose of aerial robotic navigation \cite{chenSemanticSegmentationPartially2021, geckelerLearningOccludedBranch2024a}. In all cases, the authors of those papers used, in addition to a variety of networks, RGB-D images that included depth information, rather than just RGB images. The additional feature channel provided by the depth information allows for significantly more robust segmentation when fusion between RGB and depth feature channels occurs early in the pipeline. 

\subsubsection{Scaling Issues}

A significant limitation of the method is due to the definition of a pixel-to-length ratio. As the PLI method uses only RGB data, it is not possible to accurately determine an appropriate spatial calibration between a virtual distance, measured in pixels, and the corresponding real-world distance, measured in millimetres. During the work described here the spatial calibration is informed by the assumption that a given input image is taken at an angle and distance that minimises the perspective distortion that would otherwise result in the pixel-to-length ratio being unrepresentative for the whole tree. To account for any inaccuracies arising from these assumptions, there is an uncertainty factor in the width thresholds at the analysis stage. However, a more robust solution would require the addition of depth information to the acquired images. Through the use of RGB-D data and some basic trigonometric operations, a more accurate spatial calibration could be calculated.

\subsection{Adaptation for Alternative Platforms}

Although the PLI method was developed specifically to allow a quadrotor drone equipped with a tendon-driven claw grasping device to perch on a tree branch, the methodology could be adapted and applied to any drone seeking to perch on a tree using any type of mechanism and the corresponding perching strategy. In this section, different perching strategies will be discussed and the applicability of the proposed method will be considered for each, alongside any modifications that would be required.

Firstly, T Lan et al. developed a tensile-based perching mechanism for a quadrotor drone that aims to wrap a cord of flexible material with an attached mass around a branch such that the drone can be suspended beneath the branch \cite{lanAerialTensilePerching2024a}. To be implementable with T Lan et al.'s proposed perching strategy, the current method could be modified to negate the need to factor in claw geometry when calculating viable branch widths for perching. Additionally, the method could consider a new requirement for the radius surrounding the target branch, equal to the length of the tether, to be empty. Thereby allowing the tethered mass to freely wrap itself around the target branch without becoming tangled in other branches or obstacles.

Secondly, Zheng et al. developed a metamorphic quadrotor drone capable of mid-air shape morphing \cite{zhangSeeingForestDrones2016c}. In this case, upon landing, the rear arms of the drone change shape to clamp to the tree branch below. The challenge presented here is that the perching strategy requires the drone to descend onto a tree branch from above. Therefore, there must be sufficient free space above the target perching location for the metamorphic quadrotor drone to manoeuvre. Also, as the branch will be out of sensory range while the drone performs this manoeuvre, the onboard localisation methods must be robust enough to accurately retain the relative position of the perching location site from the drone.

Thirdly, Zufferey et al. developed a perching mechanism for a flapping-wing ornithopter drone \cite{zuffereyHowOrnithoptersCan2022}. Although a less conventional drone design than a quadrotor or fixed-wing, the system still requires a method to identify a suitable perching location. In this case, the primary modification that would need to be made to the PLI method would be to consider the free space above the branch and ensure that the wings of the ornithopter are clear of any additional branches. Additionally, considering that the ornithopter must always perch dynamically and at speed, the identified tree branch will be subject to additional load as a result of the kinetic energy of the drone imparted to the branch during the perching procedure. Therefore, the minimum viable branch thickness will need to increase to compensate for the additional bending stresses generated. 

Finally, a new approach, entitled "Treecreeper Drone", is proposed by Li et al. and aims to find a solution to the issue of limited perching location availability within dense forests \cite{li2025treecreeper}. Their design aims to perch on vertical tree trunks that are easier to access as compared to horizontal tree branches. To achieve this, a bird-inspired passive perching mechanism specifically optimized for vertical tree trunks is proposed. Similar to the tail action of treecreepers (Certhiidae family) when perching on vertical tree trunks, the proposed aerial robotic platform perches on tree surfaces using a main clawed gripping mechanism combined with a supporting “tail” mechanism featuring arrays of embedded microspines.

\subsection{Applications of Perching in a Forest}

There are multiple advantages to using aerial robots capable of perching for forestry applications, compared to traditional methods.  One advantage is the manner in which a perched drone can, upon perching, operate for a greatly extended period of time as a remote sensing platform. Such a drone could support sensors for collecting micro-meteorological data or mechanisms for collecting physical samples. As an added benefit, the platform's onboard odometry can provide spatial and temporal context to the data gathered, facilitating the development of environmental models. Another advantage of a perching drone is that it can observe and select for itself when the most valuable deployment moment would be. For example, a perching drone tasked with understanding the behaviours of forest animals might perch, waiting for an animal to emerge, so that the drone can follow. Alternatively, sampling might need to be conducted at specific times, for instance taking samples immediately after rainfall.
A final advantage of using perching aerial robots is that they are considerably less invasive than alternative methods of physical sampling or deploying remote sensors. Such methods require people to climb trees, construct scaffold towers or else deploy highly invasive tower cranes or canopy cages. In terms of versatility, minimum invasiveness and cost-effectiveness, the emerging technology of perching drones could have a significant impact on forestry applications going forward.

\section{Conclusion}

\subsection{Study Summary}

In conclusion, the PLI method has proved successful in identifying an ideal perching location in a tree based on branch width, curvature and angle criteria, at an approximate rate of 76\%  (with an uncertainty of 1 data point within the sample size producing a $\pm4 \%$  uncertainty).  This result satisfies the quantified success metric as set out in section \ref{objectives}. The method used relies on a number of assumptions about the initial input data which, by using additional depth sensor data, could be eliminated to create a more robust and universally applicable method. The proposed points in the process where this data could be incorporated are shown in Figure \ref{fig:master-flowchart} (a). Furthermore, the objective of creating a methodology that can be further developed was reached by formatting and creating the graph object to store connectivity information (applications elaborated in Section \ref{future-dev}). The PLI method has been profiled to identify the areas where further improvements in code efficiency could yield future reductions in code execution time.

\subsection{Future Works}

\subsubsection{Future Experiments}

Further to the results demonstrated in this work, we believe that there is scope for additional simulations and experiments to more rigorously demonstrate the PLI method working onboard a perching drone in a real forest environment. In particular, we believe that a comparison between our PLI method, Li's PLI method and a na\"ive method that selects a random branch would be an effective validation method for our proposed solution. Additionally, evaluating these methods in both simulated and field environments would offer the opportunity to quantify the so-called "Sim-to-Real" gap for our proposed technology.

\subsubsection{Future Developments}\label{future-dev}
During the course of this work, we have identified several potential improvements and additional features, of varying complexity, that could be implemented to further develop the PLI method:
\begin{itemize}
    \item \textbf{Additional perching-relevant insight:} The PLI method as described produces the location of a potential perching site. However, the method described could be adapted to provide the drone with information that could be used for collision-avoidant path planning as well as onboard understanding of the clearance required during the approach and releases stages of perching.
    \item \textbf{Obstruction detection:} The foremost obstruction, unaccounted for in the PLI method, is the presence of dense foliage. The input dataset contains almost exclusively deciduous trees with few leaves as to allow a clearer segmentation of the tree trunk and branches. \\
    \textbf{Possible solution:} A common solution seen in related works is for the addition of a depth-channel during image acquisition. The use of RGB-D images could allow effective segmentation and removal of any foliage and an accurate interpolation between existing branches to be performed. Alternatively, assuming a given occlusion occurs only from a limited number of perspectives, the drone could circle around the medial axis of a given tree trunk until the original occlusion is no longer present.
    \item \textbf{Image Stabilisation:} As part of the branch angle determination, the PLI method assumes the input image is horizontally level. \\
    \textbf{Possible solution:} An angle offset provided by a gyroscope or other attitude sensor could be integrated to allow for this circumstance.
    \item \textbf{Branch Structural Integrity:} To prevent the risk of an identified perching branch rupturing under the perching load, especially if heavier drones are used, the PLI method should consider branch bending stress as one of the criteria.  \\
    \textbf{Possible solution:} The PLI method could incorporate tree species identification to determine the capacity of a given branch to bear the weight of the perching drone.
    \item \textbf{Dataset Suitability:} A number of suggestions to improve the segmentation dataset are outlined in Section \ref{app:dataset}.
\end{itemize}

\vspace{1cm}

{\centering The full code used during this study can be found in the GitHub repository: \\ \href{https://github.com/Leonie-G-B/Drone-Perching-CV}{Drone-Perching-CV: Leonie Bottomley (\underline{https://github.com/Leonie-G-B/Drone-Perching-CV})}.\par}

\clearpage


\appendix
\renewcommand{\thesection}{\Roman{section}} 
\section{APPENDIX I: ADDITIONAL SUPPORTING FIGURES}\label{app:imgs}

\begin{figure}[H]
    \centering
    \includegraphics[width=0.7\linewidth]{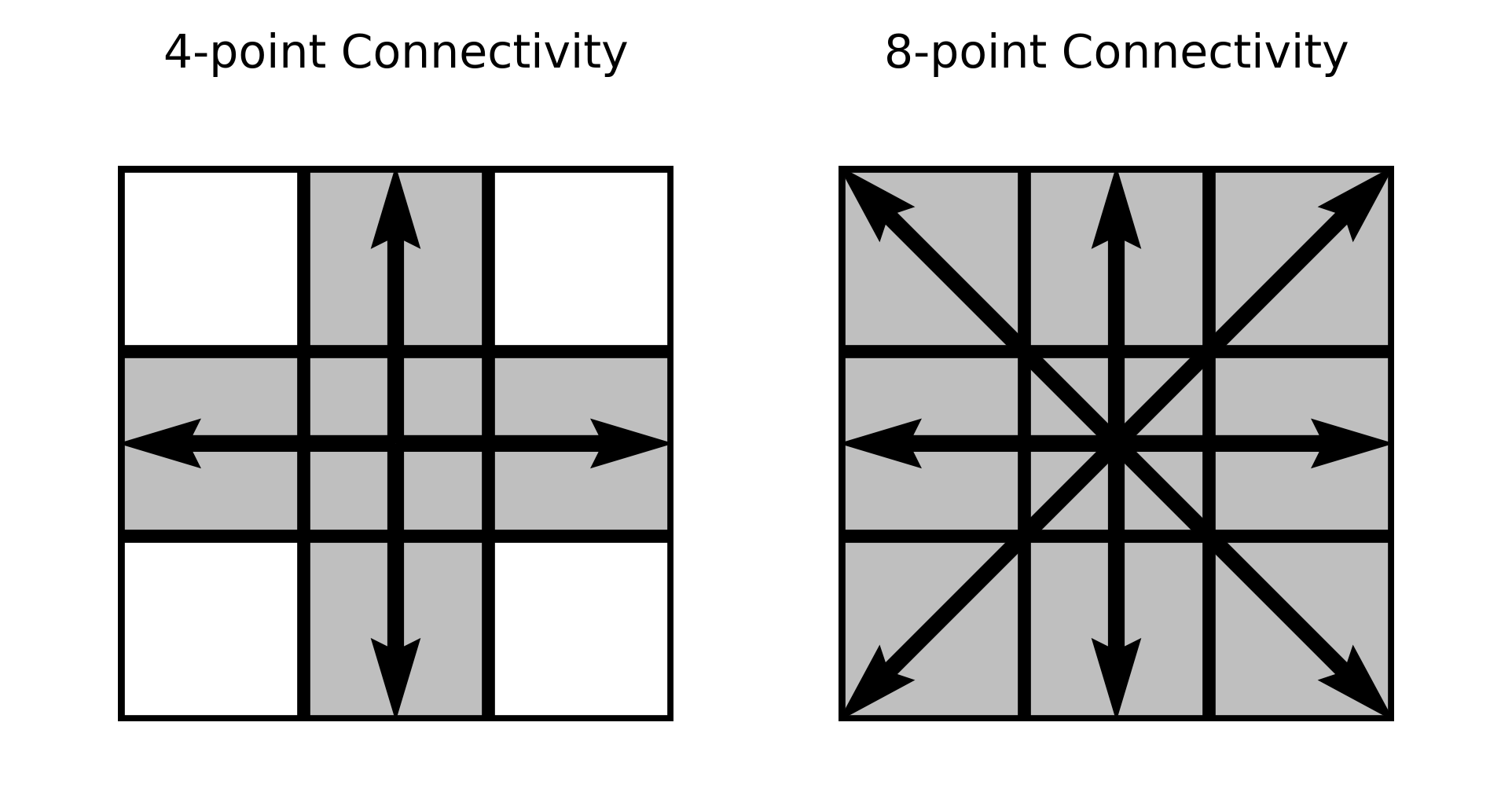}
    \caption{Visual representation of 4 and 8-point connectivity in image processing.}
    \label{fig:connectivity}
\end{figure}

\begin{figure}[H]
    \centering
    \begin{subfigure}[b]{0.45\textwidth}
        \centering
        \includegraphics[trim=18cm 2.5cm 16cm 2cm, clip, width=\textwidth]{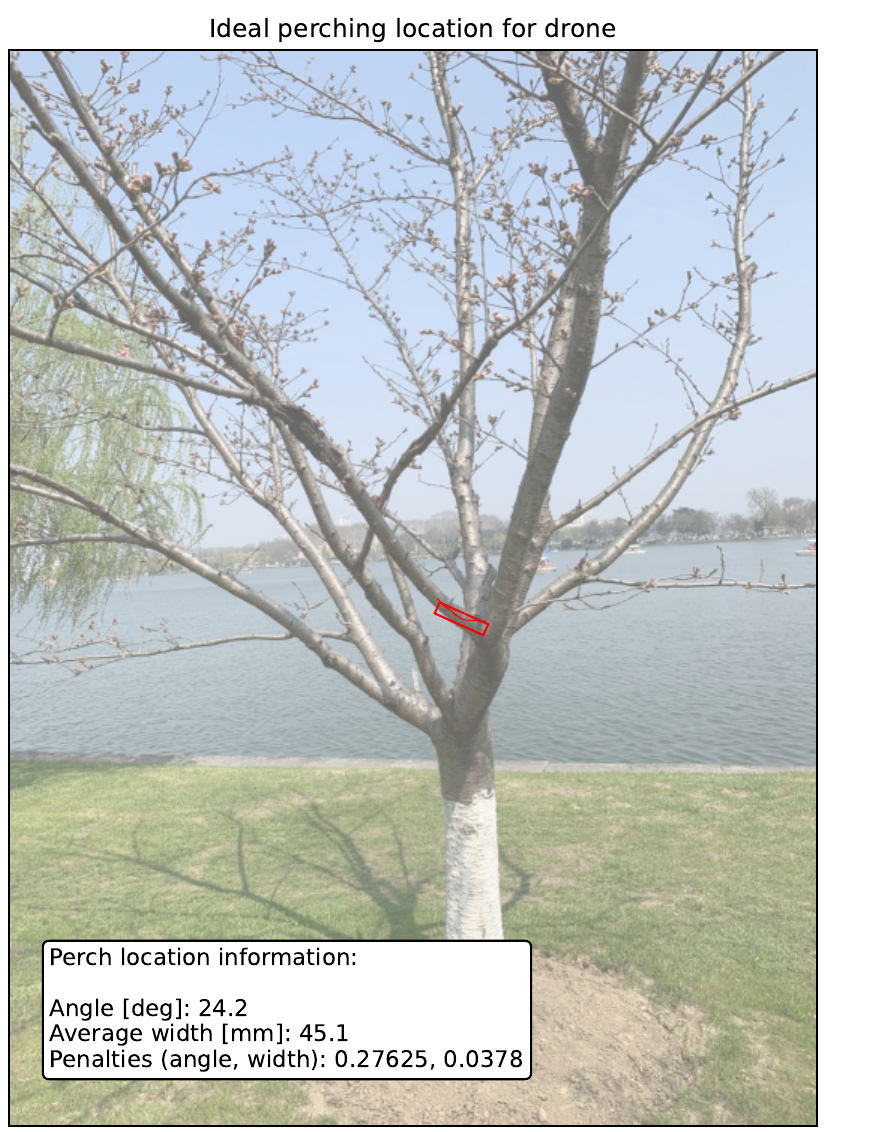}
        \caption{Perch location for $\lambda_{\text{angle}}= 0.8$ ($\lambda_{\text{width}} = 0.2$)}
        \label{fig:high-angle-bias} 
    \end{subfigure}
    \begin{subfigure}[b]{0.45\textwidth}
        \centering
        \includegraphics[trim=18cm 2.5cm 16cm 2cm, clip, width=\textwidth]{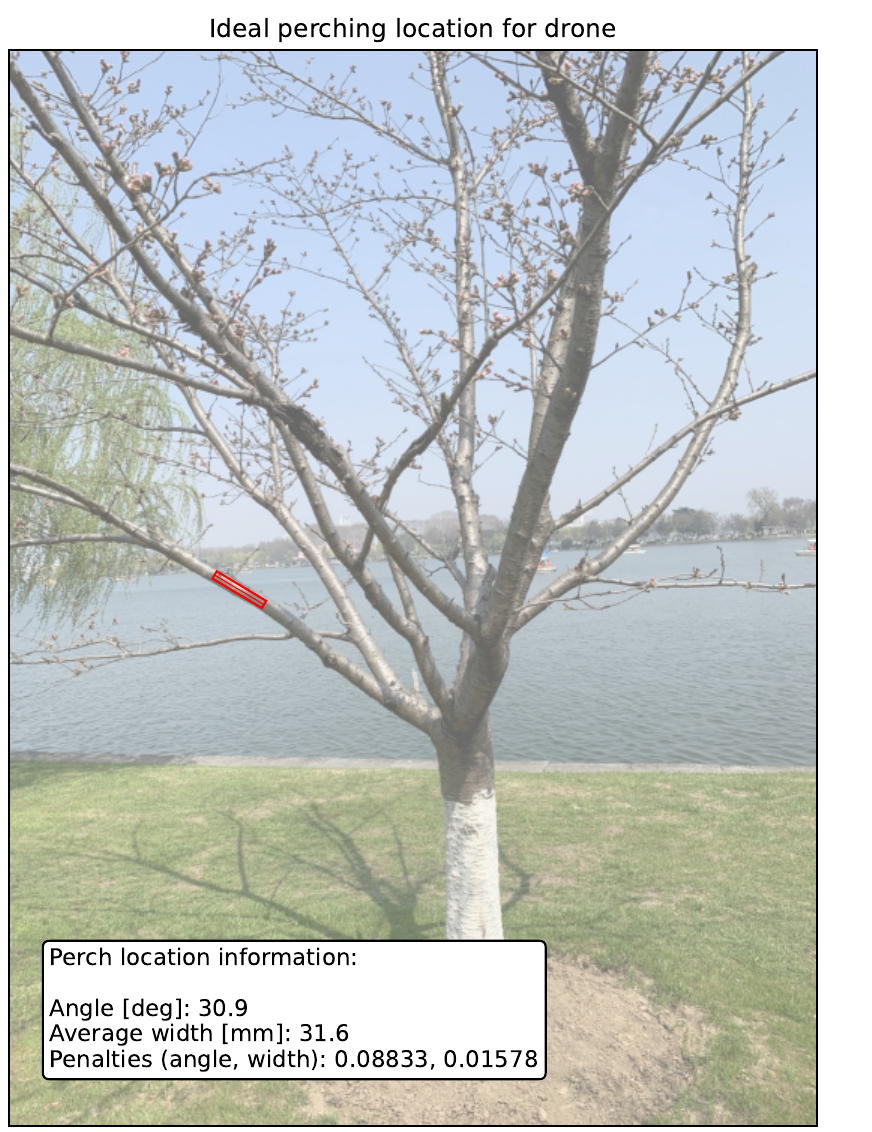}
        \caption{Perch location for $\lambda_{\text{width}}= 0.8$ ($\lambda_{\text{angle}} = 0.2$)}
        \label{fig:high-width-bias} 
    \end{subfigure}
    \caption{Perch location results for iso-input and varying penalty biases.}
    \label{fig:compare-biases}
\end{figure}

\section{APPENDIX II: FURTHER METHOD EXPLANATIONS}\label{app:more_info}

\subsection{Terminology and Techniques} \label{intro_terminology}

In order to utilise the skeletal analysis methods, the pixel-wide representation of the tree must be extracted. Two key methods were identified, both of which utilise machine learning: semantic segmentation and pose estimation.

Pre-built and trained computer vision models are readily available for all applications and performance requirements. The most labour-intensive element in applying a model to a specific use case is the creation of the input dataset. For a model to have good generalisation, a large representative training dataset is required. Performing instance segmentation for a tree (the trunk and significant branches) is not a novel challenge - there exists an ideal open-source dataset for this exact use case \cite{BranchSegmentation_2022}.

Classical image processing techniques also form the foundation of this study. Through the use of well documented and robust open-source packages such as Open-CV \cite{opencv_opencv_nodate} and Scikit-Image \cite{van_der_walt_scikit-image_2014}, key image features and fundamental structures can be extracted. The methods within these toolboxes are leveraged by the application of mathematical morphology \cite{haralick_image_1987}, which analyses image structures using operations such as erosion and dilation. These operations apply a small structuring element ('kernel') to scan the image and modify the region based on boolean conditions. As such, these techniques are optimal for application to binary images, such as the output of image segmentation models. Erosion and dilation involve the removal and addition of boundary pixels, respectively, and on this premise, this study can create a representation of the structure of a tree, which can be quantitatively analysed for perch suitability.

\subsection{Converting the Dataset for Model Training}\label{convert-dataset}

\begin{figure}[ht]
     \centering
     \begin{subfigure}[b]{0.35\textwidth}
         \centering
         \includegraphics[width=\textwidth]{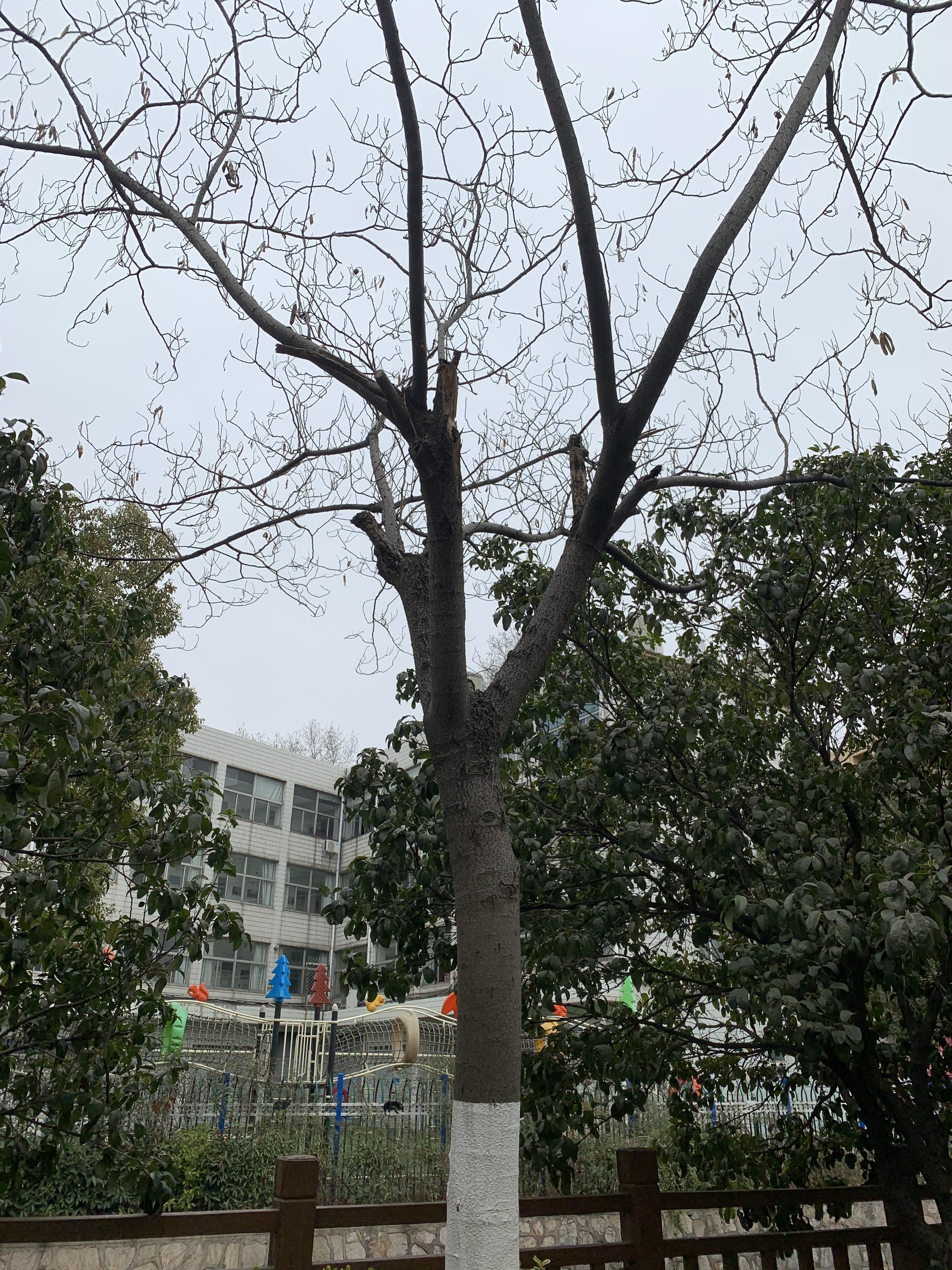}
         \caption{Original image}
         \label{fig:orig_img}
     \end{subfigure}
     \begin{subfigure}[b]{0.35\textwidth}
         \centering
         \includegraphics[width=\textwidth]{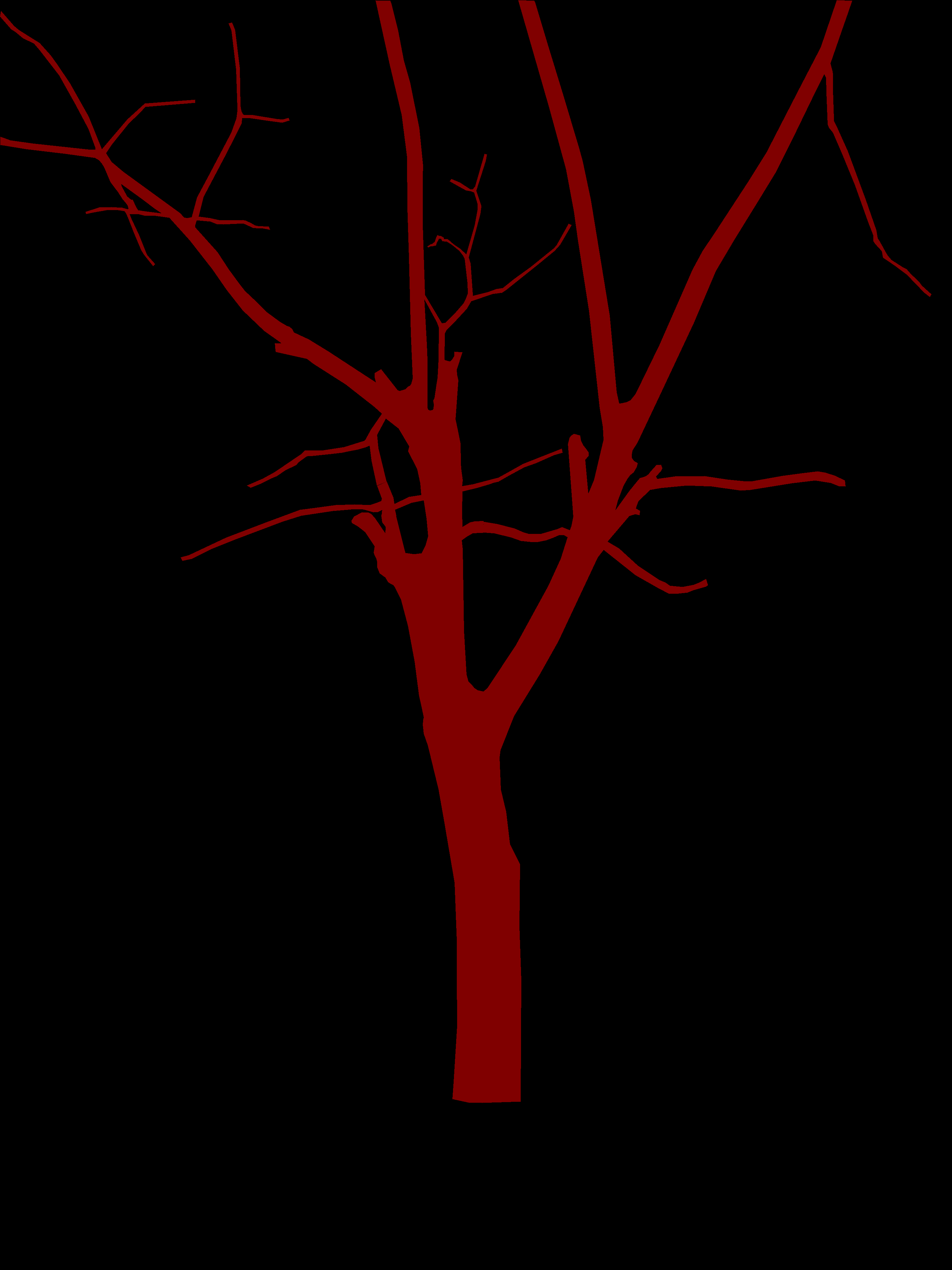}
         \caption{Segmentation mask}
         \label{fig:mask}
     \end{subfigure}
    \caption{Example image and associated segmentation mask from the used dataset.  \cite{BranchSegmentation_2022}}
    \label{figs:dataset}
    \hfill
\end{figure}

The dataset used provided the segmentation mask as a binary image (Figure \ref{fig:mask}) which is not the accepted input format for the model used. Ultralytics YOLOv11 Semantic Segmentation model requires the mask be represented as 'contours'. These are essentially co-ordinates that, when connected, form the outline of the mask. This is represented in a text file for each mask, in the structure shown in Figure \ref{fig:bounding_box_format}, where the 'class-index' allows the functionality to specify multiple segmented objects in a single image; for this study however, only one object is required and specified. 

\begin{figure}[h]
    \centering
    \begin{tcolorbox}[colback=gray!10, colframe=black, arc=3mm, boxrule=0.5pt]
        \texttt{<class-index> <x1> <y1> <x2> <y2> ... <xn> <yn>}
    \end{tcolorbox}
    \caption{YOLO Segmentation training data format.}
    \label{fig:bounding_box_format}
\end{figure}

The code repository created for this study contains a script capable of performing the pre-processing required and formatting the images into the file structure required for the model training algorithm. In order to correctly represent the mask where the contour is completely enclosed, a further processing step is required. 
The contour finding algorithm used was OpenCV's 'find\_contours'. The 'find\_contours' algorithm provides the functionality to identify 'parent' and 'child' contours, where the latter are nested within the former. To represent the mask in a way that preserves the original shape, a small 'break' in the mask is created at the thinnest point in the mask that separates a parent and child contour.

\subsection{Methods to Convert the Skeleton Branches into Ordered Vectors}\label{info:branch_ordering}

When an image-like representation of the skeleton is converted to arrays of pixel coordinates, the order of pixels is not preserved. The 'label' method, which identifies continuous objects (with 8-point connectivity), will return the coordinates in ascending y-coordinate position order. Therefore, for non-monotonic geometries that contain a reversal in the y-direction, the branch pixels will be incorrectly ordered.

Instead, a new approach was identified. The contour finding algorithm used in the dataset preparation (Appendix \ref{convert-dataset}) could also be used to identify pixel-wide objects. This algorithm proved to be universally successful, and could be applied directly to the skeleton image with no prior or post-processing required.

\clearpage


\fontsize{8}{9}\selectfont

\clearpage

\printbibliography

\end{document}